\documentclass{article} 

\usepackage[preprint]{colm2025_conference}
\usepackage{microtype}
\usepackage{hyperref}
\usepackage{url}
\usepackage{booktabs}
\usepackage{lineno}

\usepackage[normalem]{ulem}

\usepackage{graphicx}
\usepackage{tabularx}
\usepackage{soul}
\usepackage{epstopdf}
\usepackage[utf8]{inputenc}
\usepackage{svg}
\usepackage{hyperref}
\usepackage{xstring}
\usepackage{color}
\usepackage{pifont}
\usepackage{amsmath}
\usepackage{multirow}
\usepackage{xspace}
\usepackage{epsfig, endnotes}
\usepackage{array}
\usepackage{float}
\usepackage{mfirstuc, tabulary}
\usepackage{paralist}
\usepackage{enumitem}
\usepackage{wasysym}
\usepackage{amssymb}
\usepackage{pgf-pie}
\usepackage{nameref}
\usepackage{makecell}
\usepackage{csquotes}
\usepackage{floatflt}
\usepackage{wrapfig}

\usepackage{cuted}
\usepackage{xcolor}
\usepackage{bibentry}
\usepackage{url}
\usepackage{subcaption,siunitx,booktabs}

\definecolor{darkblue}{rgb}{0, 0, 0.5}
\hypersetup{colorlinks=true, citecolor=darkblue, linkcolor=darkblue, urlcolor=darkblue}

\title{\numtom{}: A Dataset for $\mathbb{D}$ynamic $\mathbb{I}$nformation $\mathbb{A}$nd $\mathbb{M}$ental modeling $\mathbb{O}$f $\mathbb{N}$umeric $\mathbb{D}$iscussions}

\author{Sayontan Ghosh$^*$, Mahnaz Koupaee$^*$, Yash Kumar Lal$^*$, Pegah Alipoormolabashi$^*$, \\ \textbf{Mohammad Saqib Hasan$^*$, Jun Seok Kang$^\dagger$, Niranjan Balasubramanian$^*$}\\
$^*$ Department of Computer Science, Stony Brook University\\
$^\dagger$ Blink Health
\\
\texttt{\{sagghosh, mkoupaee, ylal, palipoormola, mdshasan, niranjan\}}
\texttt{@cs.stonybrook.edu},\\ {jun.s.kang@gmail.com }
}

\newcommand{\numtom}{\texttt{DIAMONDs}}

\newcommand{\eat}[1]{}
\begin{document}

\ifcolmsubmission
\linenumbers
\fi

\maketitle
\begin{abstract}

Understanding multiparty conversations demands robust Theory of Mind (ToM) capabilities, including the ability to track dynamic information, manage knowledge asymmetries, and distinguish relevant information across extended exchanges. To advance ToM evaluation in such settings, we present a carefully designed scalable methodology for generating high-quality benchmark conversation-question pairs with these characteristics.  
Using this methodology, we create \numtom{}, a new conversational QA dataset covering common business, financial or other group interactions. In these goal-oriented conversations, participants often have to track certain numerical quantities (say \emph{expected profit}) of interest that can be derived from other variable quantities (like \emph{marketing expenses, expected sales, salary}, etc.), whose values also change over the course of the conversation. \numtom{} questions pose simple numerical reasoning problems over such quantities of interest (e.g., \emph{funds required for charity events, expected company profit next quarter}, etc.) in the context of the information exchanged in conversations. This allows for precisely evaluating ToM capabilities for carefully tracking and reasoning over participants' knowledge states.

%
Our evaluation of state-of-the-art language models reveals significant challenges in handling participant-centric reasoning, specifically in situations where participants have false beliefs. Models also struggle with conversations containing distractors and show limited ability to identify scenarios with insufficient information. These findings highlight current models' ToM limitations in handling real-world multi-party conversations.\footnote{Dataset and code are available at: \url{https://github.com/StonyBrookNLP/diamonds}}


\end{abstract}
\section{Introduction}





\begin{wrapfigure}{c}{0.4\textwidth}
\centering
    \includegraphics[width=0.4\textwidth]{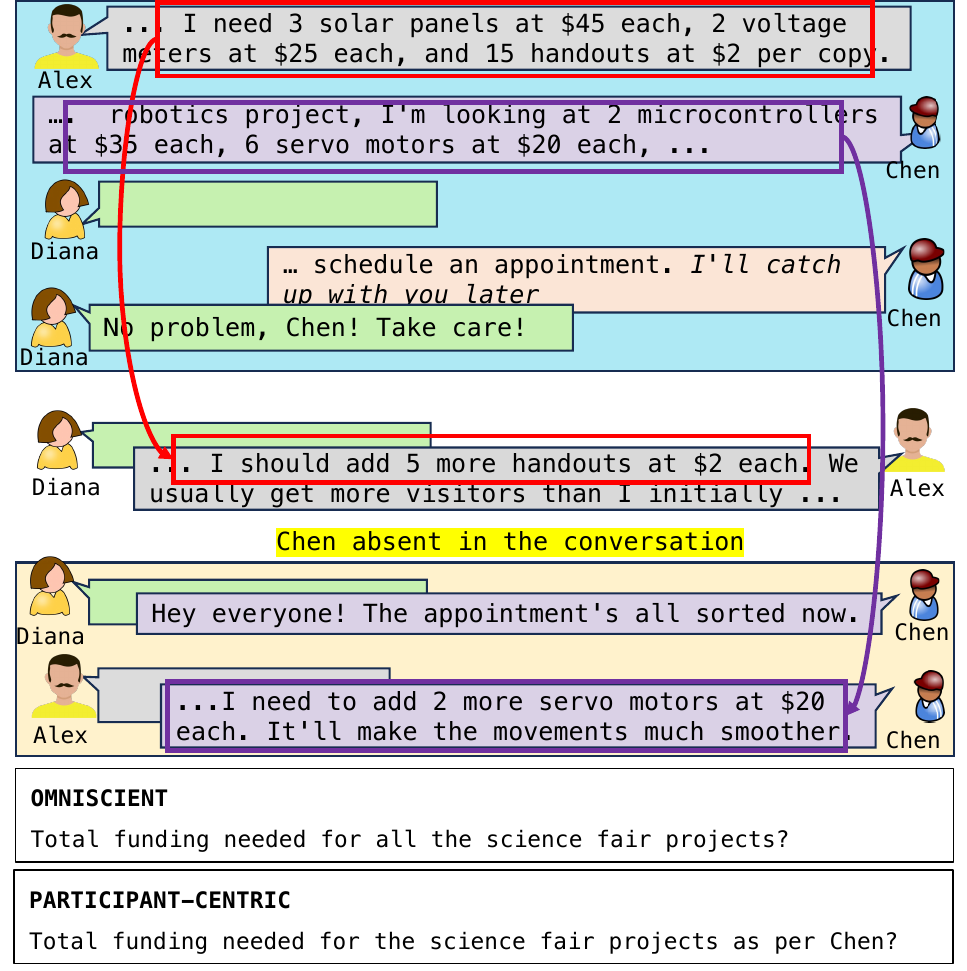}
    \caption{\small{A conversation and question example from \numtom{}. Alex increases his initial handout requirements from 15 to 20 units while Chen is absent. Chen also updates his motor requirement from $2$ to $4$ after returning.}
    }
    \vspace{-15pt}
    \label{Fig:conv}
\end{wrapfigure}

Effective communication and cooperation in multi-party conversations depends on the ability to model mental states such as beliefs, intents, desires, emotions, knowledge of oneself and others ~\citep{frith1994autism, clark1996using, pickering2004toward, de2006conversation} -- i.e., Theory of Mind (ToM) abilities. Understanding multi-party conversations thus presents a strong test bed for evaluating ToM capabilities, where participants must track not only the information being shared, but also maintain dynamic models of who has access to what information at different points in time.
This ToM challenge becomes particularly complex in information rich conversations with multiple participants, common in settings such as business meetings, financial discussions, planning sessions, etc. In these settings, the states of key variables (e.g., \emph{expenses, timelines, or task allocations} in a \emph{project budgeting discussion}) frequently change due to new information, corrections, or updates. Consider an illustrative example in Figure~\ref{Fig:conv} (\textcolor{red}{red arrow}) where during a \emph{project logistics discussion}, \underline{\emph{Alex increases his initial handout requirements from $15$ to $20$ units}} while \underline{\emph{Chen is absent}}. This seemingly simple update adds multiple reasoning demands\footnote{There are likely analogous cognitive demands for humans completing such tasks. Our goal here, however, is not draw connections with any specific set of human Theory-of-Mind capabilities but rather to design a stress test for understanding and improving models abilities to use such capabilities for conversation understanding.} for conversation understanding: tracking long-term dependencies (connecting initial estimates with later adjustments), managing dynamic information states (updating quantities as new information arrives), handling information asymmetry (Chen's outdated knowledge state leading to false belief), and distinguishing relevant details from distractors (for quantities of interest) in information-rich exchanges. 

Existing benchmarks for evaluating ToM capabilities in LLMs pose QA tasks on simple narratives, stories~\citep{sap-etal-2022-neural, gandhi2023understanding, chen-etal-2024-tombench}, or synthetic sequences of actions~\citep{le2019revisiting, wu-etal-2023-hi, xu-etal-2024-opentom}. These resources, while valuable for basic ToM assessment, are either too simplistic to capture the complexities of real-world social interactions.
Most evaluate ToM through categorical or extractive QA tasks where the context contains semantic cues directly related to the answer, avoiding the challenges of tracking and integrating information across extended exchanges. While the recently introduced FanTom dataset~\citep{kim-etal-2023-fantom} tests ToM abilities in conversations, it mainly focuses on extractive questions where relevant information is typically localized within specific dialogue segments. In contrast, real-world multi-party conversations require sophisticated ToM capabilities to track dynamic information states, manage knowledge asymmetries across multiple participants, over entire extended conversations, and integrate this understanding with other cognitive demands like numerical reasoning - challenges that existing benchmarks fail to capture.

To address these limitations, we present \numtom{}, a benchmark designed to stress-test ToM capabilities in multi-party conversations. \numtom{} focuses on tracking and reasoning over multiple dynamic variables whose states changes across the conversation, in the presence of information asymmetries between participants. We embed multi-step numerical reasoning problems within multi-party conversational contexts where successful comprehension requires both robust information tracking and modeling of participant-specific knowledge states. This ensures answer to the question is unique, not-extractive while requiring multi-step reasoning over the information across the entire conversation.

Creating numerically consistent, detailed, and highly structured conversations with specific characteristics poses significant challenges for both human annotators and LLMs. To overcome this, we implemented a multi-stage synthetic data generation process where an LLM first generates a narrative script with a corresponding question-answer pair, then transforms this script into a multi-party conversation while preserving all question-relevant information. 
Each script consists of a \emph{\textbf{premise}} that establishes the complete set of variable relevant to the question and their initial state, followed by a sequence of \emph{\textbf{variable state perturbations}} that unfold throughout the conversation and perturbs the states of these variables. For example, in Figure~\ref{Fig:conv}, the premise relevant to the question about "total funding for science fair" sets up the space of the relevant variables such as "handouts Alex needs" ($15$), "servo motors Chen needs" ($6$), their respective prices ($\$2$, $\$20$), etc., along with their initial states/values. As the conversation progresses, these initial variable states are modified through perturbations such as \emph{Chen stating that he needs $2$ more servo motors, updates his requirement from $6$ to $8$} or \emph{Alex updating his handout requirement}. This combination of initial states and subsequent variable state perturbations ensures that the state of the variables affecting the answer changes throughout the conversation, making accurate tracking of variable states essential for correct reasoning. After the LLM generates the premise and perturbation information, we integrate these elements into a coherent script using a probabilistic template sampled from a Markov process. The template contains slots for information exchanges and participant movements, defining information access patterns throughout the conversation. This approach introduces randomness into the generation process that is external to the LLM while scaffolding the premise, variable perturbation, participant movements into a coherent script.

The conversations in \numtom{} induce several cognitive challenges that stress-test ToM capabilities: (1) tracking variable states as new information emerges, (2) resolving long-term dependencies between information shared at different points, (3) maintaining separate knowledge models for each participant based on their presence during information exchanges, (4) distinguishing relevant information from distractors, and (5) identifying information gaps. By presenting both omniscient questions (testing global understanding) and participant-centric questions (testing the ability to reason from specific participants' perspectives, including their potentially false beliefs), \numtom{} provides a comprehensive evaluation of ToM capabilities in complex multi-party settings.

Our evaluation of state-of-the-art LLMs (\texttt{gpt-4o}~\citep{achiam2023gpt}, \texttt{Claude-3.5-Sonnet}, and \texttt{Llama3}~\citep{dubey2024llama}) on \numtom{} reveals significant gaps in their ToM capability in context of multi-party conversation. Benchmarking reveals significant challenges in handling participant-centric reasoning, with performance dropping from $80.0\%$ on omniscient questions to $55.1\%$ on participant-centric questions. This performance further drops to $27.0\%$ in participant-centric questions for participants with false beliefs. Models also show decreased performance when dealing with conversations with distractors and limited ability to identify scenarios with insufficient information. These findings highlight the limitations in ToM capability of current models needed in multi-party conversations.
\section{Related Works}
\vspace{-5pt}
\paragraph{2.1 Conversation Comprehension:}
Existing conversation comprehension research has strong focus on classification tasks like slot filling (MultiWOZ~\citep{10.1007/978-3-030-88483-3_16, ye-etal-2022-multiwoz}),  intent classification (CLINC150~\citep{larson-etal-2019-evaluation}, SILICONE~\citep{chapuis-etal-2020-hierarchical}), and emotion detection (MELD~\cite{poria-etal-2019-meld}, EmoWOZ~\citep{feng-etal-2022-emowoz}). These tasks do not require complex multi-step reasoning capabilities across conversation that our work requires.
Dialogue summarization represents another major research direction, with datasets like SAMSum~\citep{gliwa-etal-2019-samsum}, MeetingBank~\citep{hu-etal-2023-meetingbank}, DialogSum~\citep{chen-etal-2021-dialogsum}, and MediaSum~\citep{zhu-etal-2021-mediasum}. While valuable, these tasks are primarily extractive and don't demand the precise tracking of dynamic numerical information central to our framework.
Conversation-based QA has advanced through datasets such as CoQA~\citep{reddy2019coqa}, QuAC~\citep{choi2018quac}, ShARC~\citep{saeidi-etal-2018-interpretation}, OR-QuAC~\citep{10.1145/3397271.3401110}, and QReCC~\citep{anantha-etal-2021-open}. The most relevant work, Fantom, evaluates theory-of-mind capabilities in conversations with information asymmetry. However, Fantom~\citep{kim-etal-2023-fantom} primarily tests extractive fact-based questions localized within conversation segments, whereas our work requires non-extractive, multi-step mathematical reasoning across multiple dynamic variables whose states evolves throughout entire conversations.

\paragraph{2.2 Theory of Mind Benchmarks:} Narrative-based ToM benchmarks evaluate mental state reasoning through narrative contexts. Recent works include ToMi~\citep{le-etal-2019-revisiting}, which requires tracking belief states across multiple characters; Neural ToM~\citep{sap-etal-2022-neural}, which tests the understanding of beliefs and intentions; HiToM~\citep{wu2023hitom}, which examines higher-order ToM; and OpenToM~\citep{xu-etal-2024-opentom} introduces the dimension of participant character in the narrative context. While valuable for basic ToM assessment, these benchmarks often use simplified narratives with clear semantic cues, making them susceptible to reporting bias and lacking the dynamic complexity of real-world interactions.
Conversation-based benchmarks better approximate real-world social interactions by evaluating ToM within dialogue contexts. \cite{soubki-etal-2024-views} introduced the first ToM resource on real human dialogues, while FaNToM~\citep{kim-etal-2023-fantom} introduced synthetically generated conversations with information asymmetry, and NegotiationToM~\citep{Chan2024NegotiationToMAB} extended evaluation to strategic interactions. However, these benchmarks typically focus on extractive or classification tasks where answers appear in localized dialogue segments. In contrast, \numtom{} features conversations requiring multi-step reasoning over dynamic variables that evolve throughout extended exchanges, better reflecting the cognitive demands of real-world multiparty interactions.

\section{\numtom{}: $\mathbb{D}$ynamic $\mathbb{I}$nformation And $\mathbb{M}$ental modeling $\mathbb{O}$f
$\mathbb{N}$umeric $\mathbb{D}$iscussions}
\paragraph{3.1 Overview:}

\numtom{} is a conversation QA dataset comprising of $3786$ $(C, q, a)$ triples of multiparty conversation $C$, a question $q$ to answered based on $C$, and its answer $a$. 
The dataset evaluates Theory of Mind (ToM) capabilities in information-rich discussions, where values of quantities (variables) affecting  the answer to the question, evolve over the conversation discourse. 
%
Each conversation in \numtom{} follows a common structure: (1) early exchanges establish the variable space and their initial states (2) subsequent interactions introduce modifications to these states (3) participant movement patterns create information asymmetries. \numtom{} includes both omniscient (testing global understanding) and participant-centric questions (requiring reasoning from specific perspectives, including false beliefs). It also contains conversation variants with thematically relevant distractors and deliberately underspecified scenarios resulting in unanswerable questions. All questions are non-extractive, requiring multi-step numerical reasoning over the entire conversation.

\paragraph{3.2 Dataset Creation}
\begin{figure*}[h!]
\vspace{-10pt}
    \centering
    \includegraphics[width=\textwidth]{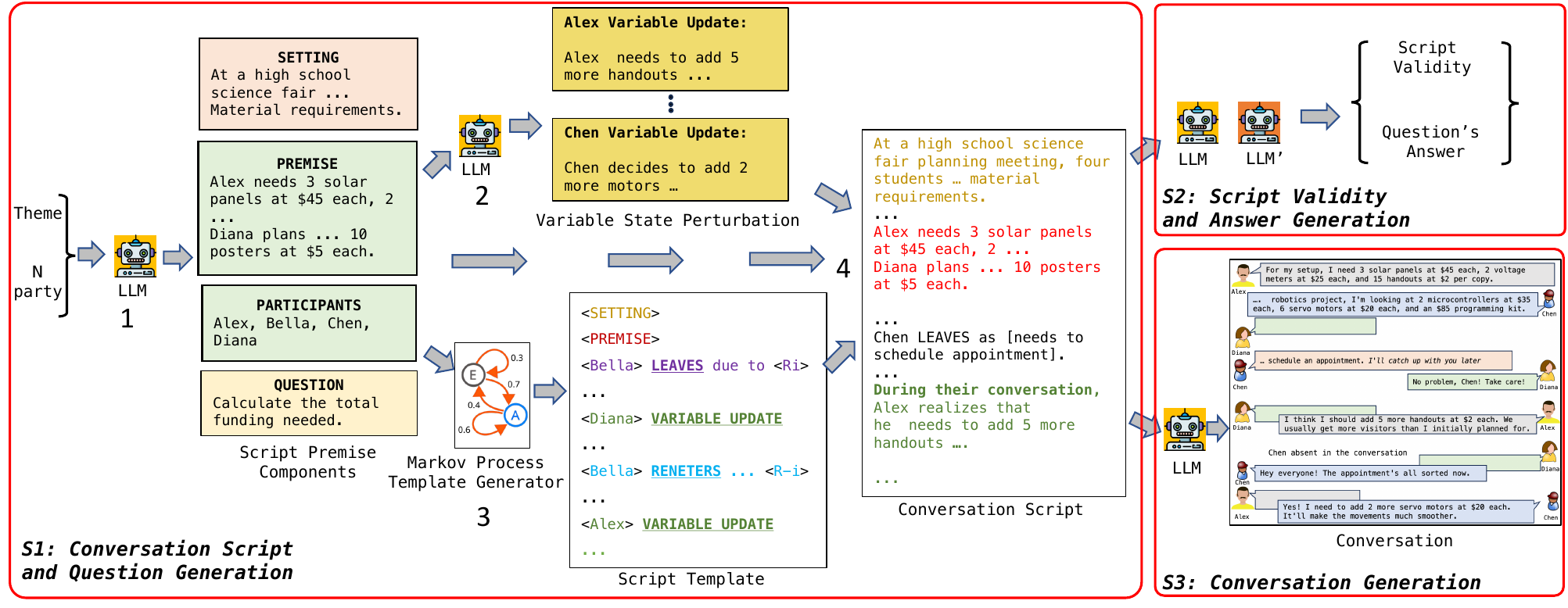}
    \caption{\small{\textbf{Dataset Generation}: \emph{\textbf{S1}}: Script generation begins by (1) creating premise components (setting, premise, question, participants) based on a \textbf{\emph{theme}} $t$ and $n$ \textbf{\emph{participants}} (2) generating \textbf{\emph{variable state perturbations}} that modify initial conditions, (3) producing a \textbf{\emph{script template}} via a Markov process (4) assembling these elements into a complete \textbf{\emph{script}}. \emph{\textbf{S2}}: Script validation - evaluates logical consistency and generates verified answers. \emph{\textbf{S3}}: Conversation translation - transforms the validated \textbf{\emph{script}} into a natural \textbf{\emph{multi-party conversation}} while preserving all critical numerical information.}} 
    \label{Fig:conv_gener_pipeline}
\end{figure*}

We target automatic generation of conversations following works that uses LLMs~\citep{kim-etal-2023-fantom,xu-etal-2024-opentom} and templates~\citep{gandhi2023understanding} for the generation of conversations and narratives. Our primary design goals for \numtom{} are to include problems that: (i) are non-extractive, i.e., resistant to simple localized extraction, (ii) require tracking and aggregating information across the entire conversation, (iii) have unambiguous answers enabling precise evaluation, and (iv) include information asymmetry. To this end, we want to obtain conversations with multiple thematically relevant numeric quantities in a consistent manner, include information asymmetries, and be coherent overall. 

This, in itself, poses a challenging generation problem.  From our initial attempts, we found that directly generating such structurally constrained conversations using few-shot prompting produced inconsistent and incoherent results. For example, LLMs frequently failed to maintain numerical consistency across long exchanges or struggled to properly implement the required information asymmetries.
To address this, we leverage the strong instruction following and refinement capabilities of LLMs to devise a $3$-stage pipeline as shown in Figure~\ref{Fig:conv_gener_pipeline}. The key idea is to use a \emph{script} as a basis for generating the \emph{conversation} and it QA pairs. The \emph{script} describes how the numerical quantities will be introduced and modified in the \emph{conversation} and how the information asymmetries will arise. Also, generating correct QA pairs based on the script is easier and allows for effective validation.

\eat{\begin{itemize}[leftmargin=1cm]
    \setlength\itemsep{-0.3em}
    \item[\emph{\textbf{S1}}] \textbf{Script and Question Generation} - For a given theme $t$ and $n$ participants, we generate a script $S$ and a question $q$ based on it.
    \item[\emph{\textbf{S2}}] \textbf{Script Validation and Answer Derivation} - We validate the logical consistency of $S$ and derive the solution to $q$ based on it.
    \item[\emph{\textbf{S3}}] \textbf{Conversation Creation} - We transform the validated script $S$ into a natural multiparty conversation $C$ while preserving all numerical information and participant dynamics.
\end{itemize}
}
\paragraph{S1: Generate conversation script and question: }
A conversation \emph{script} $S$ is a structured narrative that captures essential interactions between participants ($P_1, \cdots, P_n$). Each script consists of two key components: (i) a \emph{\textbf{premise}} segment establishing the initial scenario and variable states, and (ii) subsequent segments introducing \emph{\textbf{variable state perturbations}} through participant interactions (e.g., \emph{Chen adds two motors}), with controlled participant entry/exit events creating information asymmetry (e.g., \emph{Chen leaves in the middle for an appointment}). To create diverse and realistic patterns of participant engagement, we model various ways participants enter, exit, and rejoin conversations.
We model these dynamic participation patterns using a Markov process that generates \emph{\textbf{script templates}} with different information access configurations. As shown in Figure~\ref{Fig:conv_gener_pipeline}, the process can be used to sample templates with designated slots for critical information exchanges and participant movements. A snippet of a sampled template is show below:
\vspace{-6pt}
\begin{quote}
\small
$\cdots$
$\rightarrow$
$[P_i]$ \texttt{LEAVES DUE TO} $[R_m]$
$\rightarrow$
\texttt{VARIABLE UPDATE BY} $[P_j]$
$\rightarrow$
$[P_k]$ \texttt{RETURNS}.
$\rightarrow \cdots$ \\
\vspace{-6pt}
$\cdots$
$\rightarrow$
$[P_k]$ \texttt{LEAVES DUE TO} $[R_m]$
$\rightarrow$
\texttt{CASUAL TALK}
$\rightarrow$
$[P_k]$ \texttt{RETURNS}
$\rightarrow\cdots$
\end{quote}
As a final \emph{\textbf{assembling}} step, these templates are then populated with relevant content and to form the complete script. The following subsections detail this process.

\paragraph{1. Script Premise Generation}
Given the number of participants $n$ and a conversation theme $t$, we generate the premise components that establish the conversation context and initial variable states. This includes: (i) \textbf{setting} - A concise context description for the conversation (ii) \textbf{Script Premise} - a detailed math word problem describing the conversation involving numerical information exchanged with each of them. (iii) \textbf{question} - A question about a numeric quantity that can be derived from the information in the script premise (iv) \textbf{participants} - The list of conversation participants.
We generate the script premise components using few-shot prompting with a LLM. Appendix, Figure~\ref{Fig:premise_dict} shows these components for a \emph{science fair planning} scenario with $4$ participants, with the prompt in Figure~\ref{Fig:premise_dict}.

\paragraph{2. Variable State Perturbation Generation}
The numeric quantities (\emph{variables}) of interest can change values as the conversation unfolds (e.g. the \emph{material requirements} keep changing). We model these changes by generating \emph{perturbations} that alter the values of the \emph{variables} associated with each participant. The perturbations are added only for \emph{variables} that are relevant to answering the question. For instance, in Figure~\ref{Fig:conv}, perturbation for \emph{Chen} (\emph{he needs to add $2$ more motors}), results in increasing the number of motors he requires from $6 \rightarrow 8$ . This change in turn increases the total cost for Chen and the overall \emph{funding needed for the project}, changing the answer to the question. 
To maintain logical consistency and manage complexity, we implement two key design constraints: (i) we only perturb independent variables from the premise, avoiding modifications to variables constrained by multiple relationships that could create logical inconsistencies (ii) we ensure each \emph{perturbations} is expressed independently rather than in terms of other \emph{perturbations}, removing ordering dependencies when arranging them in the conversation.
We generate these \emph{variable state perturbations} using LLM with few-shot prompting ( prompt shown in Appendix, Figure~\ref{Fig:var_ch_prompt}).

\begin{wrapfigure}{c}{0.5\textwidth}
\includegraphics[scale=0.4]{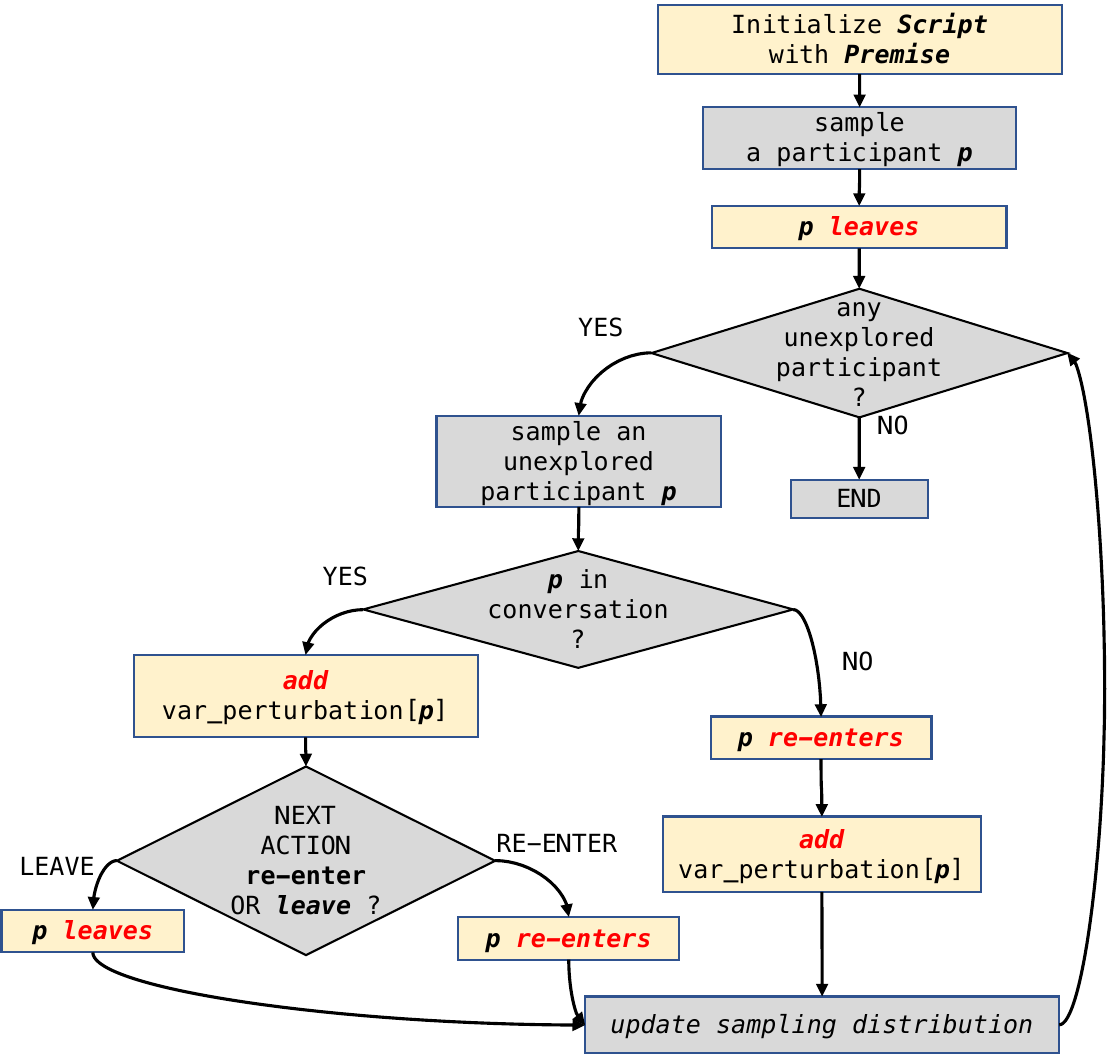}
\caption{\small{A simplified Markov process generating script \emph{template} with slots for \emph{script premise}, \emph{variable state perturbations}, and participant movements.}}
\vspace{-40pt}
    \label{Fig:script_tmplt_grmmr0}
\end{wrapfigure}

\paragraph{3. Script Template Generation}
We generate probabilistic script \emph{templates} from a Markov process (Figure~\ref{Fig:script_tmplt_grmmr0}) that models participant movement and information flow. This \emph{template} contains slots for \emph{premise} information, \emph{variable-state perturbations}, casual dialogue cues, and participant entry/exit indicators. The structure ensures discourse coherence while creating controlled information asymmetry, allowing us to precisely track which participant has access to what information throughout the conversation.
As shown in Figure~\ref{Fig:script_tmplt_grmmr0}, the Markov process begins with \emph{premise} slot for the \emph{script premise} with all participants intially present. It then systematically models participant movement dynamics by: (1) randomly selecting participants to leave with plausible reasons, (2) determining which participant shares new information (\emph{variable state perturbations}) next, and (3) managing re-entries of absent participants. 
To maintain naturalistic conversation flow, we reduce the probability of repeated exits by the same participant and include casual conversation segments between major events. 
Figure~\ref{Fig:conv_template} shows a script template for $4$ participants, with full details of the template generation process in Appendix~\ref{appx:script_tmplt_grmmr}.
\paragraph{4. Assembling the Script}
The final conversation \emph{script} is assembled by combining the \emph{template slots} with information in the \emph{script premise components} and \emph{variable state perturbations}. Figure~\ref{Fig:full_script} shows the complete \emph{script} after filling the template slots (Figure~\ref{Fig:conv_template}) with information in \emph{Script Premise Components} (Figure~\ref{Fig:premise_dict}) and its \emph{variable state perturbations} (Figure~\ref{Fig:var_change}).
\paragraph{Generating Underspecified and Distractor Script Variants:}
To evaluate model robustness, we create two controlled script variants:
\paragraph{a. Distractor Variants:} We augment base scripts with thematically relevant but question-irrelevant information by introducing new quantities in the \emph{premise} and their state perturbations in one of the subsequence segment. These additions maintain conversational coherence while being carefully having no impact on the answer to the question, testing models' ability to filter relevant from irrelevant information.
\vspace{-10pt}
\paragraph{b. Underspecified Variants:} We create underspecified scripts by systematically removing critical information needed to answer questions. This is done by either (1) eliminating key details from the premise or (2) making variable state perturbations ambiguous. These scripts test a model's ability to recognize when information is insufficient rather than forcing an incorrect answer. The details for both are described in Appendix, subsection~\ref{subsec:under_distr_descr}

\subsubsection*{S2: Script Validation and Answer Generation}\label{subsec:script_val}
\vspace{-10pt}
\begin{figure*}[h!]
    \centering
    \includegraphics[width=\textwidth]{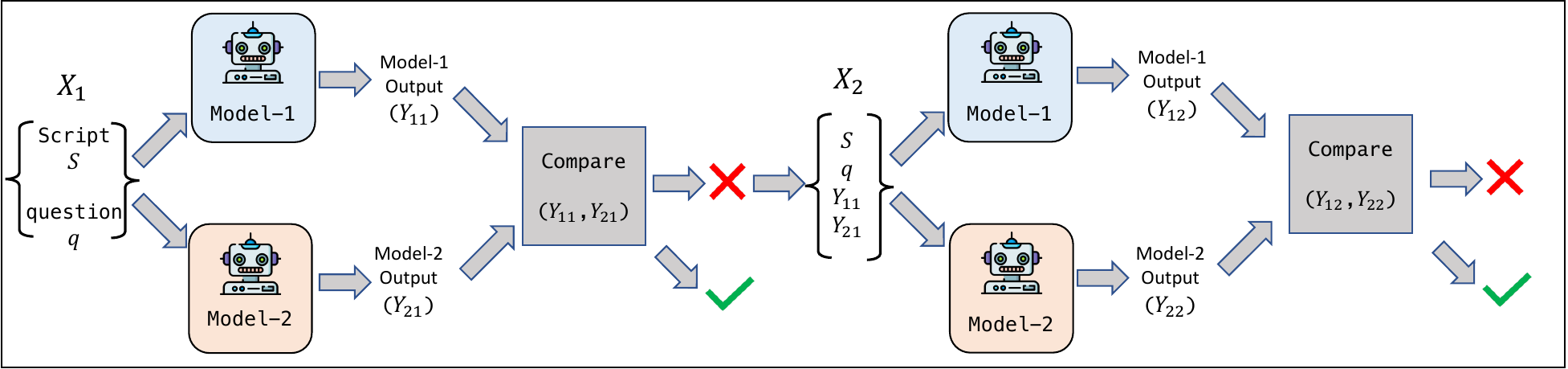}
\caption{\small{\textbf{Script validation and solution generation}: Two LLMs independently evaluate inputs and generate solutions in the first round. If their solutions match, the solution is accepted. Otherwise, a second round incorporates both first-round outputs for refinement. Agreement in the second round leads to acceptance, otherwise the input is rejected.}}
    \label{Fig:val_script}
\end{figure*}

We design a two-step reflect-refine inter-model consistency \citep{chowdhury-etal-2024-cross, AmiriMargavi2024EnhancingAR, zhang-etal-2024-self-contrast} based approach to generate solution for the question based on script. As shown in Figure~\ref{Fig:val_script}, two independent LLMs generate solutions for each script-question pair. Each solution includes evaluations of script logical consistency, question answerability, step-by-step reasoning, and a final numeric answer. If both models produce equivalent outputs on the first attempt, we accept the script and solution. When disagreements occur, we conduct a second evaluation round where both models review the script alongside both the first-round solutions. Agreement in this second round leads to acceptance, while continued disagreement results in the script being discarded.

\paragraph{S3: Conversation Generation}

We translate the script into conversation with few-shot prompting and 2-step self-reflection and refinement~\citep{madaan2024self}. The task instruction has two key focuses: (i) conversation completely covers all information in the script (ii) Avoiding any extra information not in the script that might change the answer to the question.
The prompt also includes additional guidance to maintain conversational coherence and naturalness. The prompt used for this step is shown in the Appendix, Figure~\ref{Fig:script2conv_prompt}. 


Using this process we created, \numtom{}, which contains $584$ conversations with $4$ to $7$ participants, with $3786$ unique conversation-question pairs covering $49$ conversation themes.

\paragraph{3.1 Data Quality Assessment:} 
We conducted rigorous quality evaluations on our dataset of script-conversation-question-answer tuples $\{(S, q, a, C)\}$. Our assessment focused on two critical aspects: (i) script-answer validity of $(S, q, a)$ (ii) script-conversation alignment between $(S, C)$. We randomly selected $66$ samples for evaluation by $5$ expert annotators (STEM graduate students), with each instance assessed by $2$ evaluators.\\
\textbf{(i) Script-based Answer Validity}
To determine the validity of a $\{(S, q, a)\}$ instance, the evaluators assessed: 
(a) The script $S$ was logically consistent
(b) The question $q$ was answerable based on $S$ 
(c) The solution $a$ was correct for $q$ in context of $S$. 
Of evaluated instances, $91.1\%$ of $(S, q, a)$ tuples were assessed as valid.\\
\textbf{(ii) Script-Conversation Alignment}
Evaluators evaluated the alignment of a conversation $C$ with its script $S$ by verifying that: (i) All question-relevant script details were preserved in the conversation 
(a) Conversation segments maintained the discourse order established in the script
(b) No additional calculations beyond those in the script were introduced
(c) Participant movements and information states remained consistent
Of the evaluated $(S, C)$instances, $100\%$ of the conversations were correctly aligned with their scripts.

Based on these evaluations, we estimate the validity of our curated dataset to be $\sim 91\%$.

\vspace{-10pt}
\section{Benchmarking Conversation Comprehension with \numtom{}}
\vspace{-5pt}
\paragraph{Conversation Comprehension Tasks:}
%
We introduce a QA task that requires Theory-of-Mind (ToM) abilities for conversation comprehension. This involves tracking, and reasoning with multiple dynamic numeric variables over a long multi-party conversation. Our task covers two settings: (i) \textbf{Omniscient Questions}: These require answering based on the entire conversation, reflecting complete knowledge of all exchanged information.
(ii) \textbf{Participant-Centric Questions}: These require answering from a specific participant's perspective, using only information exchanged in the conversation that participant had access to. Here participants may miss critical updates relevant to the question due to being absent from the conversation at that time. This leads to a \textbf{false belief} situation where the answer to the question from their perspective would be different from the true answer to the question.
\vspace{-5pt}
\paragraph{Evaluation Metrics:} Since the task requires numeric computations, we allow a small tolerance for minor computational or rounding errors. For base and distractor conversations, we consider a model's answer correct if its final numerical response is within $2\%$ of the true value, accounting for potential intermediate rounding errors. For underspecified conversations, a response is deemed correct if it successfully identifies the question as unanswerable.
To ensure uniform comparison, we evaluate models on a common subset of data where all models provided valid responses. This approach was necessary because models occasionally failed to generate parseable responses due to invalid JSON formatting or other runtime API errors
. Results on the full set of valid responses are in Appendix~\ref{appx:full_result}. Models show same relative performance, and the difference on overall accuracy (complete set vs common subset) ranges from $\sim0.4\%$ for \texttt{gpt-4o} to $\sim2.5\%$ for \texttt{Llama-90B}.
\vspace{-5pt}
\paragraph{Models: }
Our goal is to benchmark the inherent ToM capabilities of large language models and thus do not consider fine-tuning models on this task.
Therefore, we benchmark three state-of-art (i) closed models: \texttt{Claude-3-5-sonnet-20241022}, \texttt{gpt-4o 2024-08-06} (ii) open models: \texttt{llama3-1 405B-instruct} , \texttt{llama3-2 90B-instruct}, and \texttt{llama3-2 11B-instruct} on the task with zero-shot prompting with chain-of-thought and 5-rounds of self-reflect and self-refinement~\citep{madaan2024self}. Given the objective nature of the task, we set the generation temperature to $0$ for benchmarking. Dataset generation details are mentioned in Appendix, subsection~\ref{subsec:dataset_genr}. All the prompts used for data generation are shown in subsection~\ref{subsec:prompts}
%
\vspace{-5pt}
\section{Results}
\subsection{Participant-Centric Questions are harder than Omniscient Questions}
\renewcommand{\arraystretch}{1.1}
\begin{table*}[h!]
\small
\centering
\begin{tabularx}{\textwidth}{l p{0.16\textwidth} p{0.22\textwidth} p{0.16\textwidth}}
\toprule
\multicolumn{4}{c}{\texttt{MODEL ACCURACY ($\%$) ON BASE/REGULAR CONVERSATIONs}} \\
\toprule
\textbf{Model} & 
{\textbf{Omniscient}} & 
{\textbf{Participant Centric}} & 
{\textbf{Overall}}\\
\midrule
\texttt{claude-3-5-sonnet-20241022} & $80.0$ & $55.1\%$ & $59.6\%$ \\
\texttt{gpt-4o-2024-08-06} & $73.8$ & $39.2\%$ & $45.6\%$ \\
\texttt{llama3-1-405b-instruct} & $69.2\%$ & $37.9\%$ & $43.6\%$ \\
\texttt{llama3-2-90b-instruct} & $50.0\%$ & $24.3\%$ & $29.0\%$ \\
\texttt{llama3-2-11b-instruct} & $11.5\%$ & $9.5\%$ & $9.8\%$ \\
\bottomrule
\end{tabularx}
\caption{\small{\textbf{Omniscient} is the $\%$ of correctly answered omniscient questions, \textbf{Participant Centric} is the $\%$ of correctly answered participant-centric ones, and \textbf{Overall} is the accuracy across all the question type.}}
\label{tab:base_conv_perf}
\vspace{-10pt}
\end{table*}
%
Across all models, there is a consistent performance gap between omniscient and participant-centric questions. For base conversations (Table~\ref{tab:base_conv_perf}), \texttt{Claude-3.5} has the highest performance with $80.0\%$ accuracy on omniscient questions but drops to $55.1\%$ on participant-centric ones. \texttt{gpt-4o} shows a similar pattern with $73.8\%$ and $39.2\%$ for omniscient and participant-centric respectively. The \texttt{llama3} models follow similar trend trend, with performance scaling with model size - from the $405$B parameter version ($69.2\%$ omniscient, $37.9\% $participant-centric) down to the $11$B version ($11.5\%$ omniscient, $9.5\%$ participant-centric). This substantial performance gap can be attributed to the added complexity of modeling information access in situations with information asymmetry, compared to using the entire context. These models struggle to ignore contextually present but inaccessible information, highlighting the limitations of their ToM capabilities for multi-party conversations.

\subsection{Participant-Centric Performance: True vs False Belief}
\vspace{-10pt}
\renewcommand{\arraystretch}{1.1}
\begin{table*}[h!]
\small
\centering
\begin{tabularx}{\textwidth}{l p{0.16\textwidth} p{0.16\textwidth} p{0.22\textwidth}}
\toprule
\multicolumn{4}{c}{\texttt{MODEL ACCURACY ($\%$) ON PARTICIPANT-CENTRIC QUESTIONS IN BASE CONVERSATION}} \\
\toprule
\textbf{Model} & 
{\textbf{True Belief }} & 
{\textbf{False Belief}} & 
{\textbf{Overall}}\\
\midrule
\texttt{claude-3-5-sonnet-20241022} & $78.4$ & $28.1$ & $55.1$ \\
\texttt{gpt-4o-2024-08-06} & $54.8$ & $21.0$ & $39.2$ \\
\texttt{llama3-1-405b-instruct} & $51.6$ & $21.7$ & $37.9$ \\
\texttt{llama3-2-90b-instruct} & $31.2$ & $16.1$ & $24.3$ \\
\texttt{llama3-2-11b-instruct} & $10.5$ & $8.6$ & $9.5$ \\
\bottomrule
\end{tabularx}
\caption{\small{Model Performance on participant-centric questions in \numtom{}'s base conversations. \textbf{True Belief} is the accuracy for cases where participants have access to all the question-relevant information, \textbf{False Belief} represents cases where participants miss critical updates, resulting in outdated or incorrect knowledge states. \textbf{Participant Centric} is the overall performance across both categories.}}
\label{tab:base_participant_conv_perf}
\end{table*}

Table~\ref{tab:base_participant_conv_perf} breaks down the model performance on participant-centric question. We distinguish between True Belief scenarios, where participants have access to the correct information needed to answer questions accurately, and False Belief cases, where participants miss critical information updates, leading to a mismatch between their knowledge state and reality. This analysis reveals two critical insights:
\emph{First}, all models show dramatic performance drop when handling false beliefs situations. \texttt{Claude-3.5} achieves $78.4\%$ accuracy with true beliefs but only $28.1\%$ with false beliefs. Similar patterns appear across all models, with \texttt{gpt-4o} dropping $33.8\%$ points and \texttt{llama3-405B} declining $29.9\%$.
\emph{Second}, even for true belief participant-centric questions, most models perform worse than on omniscient questions ( Table~\ref{tab:base_conv_perf}). Since participants with true belief have access to correct information, performance should theoretically be comparable to omniscient question performance. While \texttt{Claude-3.5} shows only a small gap of $1.6\%$, other models exhibit larger differences: \texttt{gpt-4o} drops $19\%$ and \texttt{llama3-405B} falls $17.6\%$ compared to omniscient questions.
These findings highlight models' limitations in Theory of Mind capabilities as they struggle noticeably when reasoning from perspectives with incomplete information and show degraded performance even when merely adopting a participant's viewpoint rather than an omniscient one.


\subsection{Impact of Distractors on Model Performance}
\renewcommand{\arraystretch}{1.1}
\begin{table*}[h!]
\small
\centering
\begin{tabular}{@{} l | ccc | c @{}}
\toprule
\multicolumn{5}{c}{\texttt{MODEL ACCURACY (\%) ON CONVERSATIONS}} \\
\midrule
& \multicolumn{3}{c|}{\texttt{\textbf{DISTRACTOR}}} & 
\multicolumn{1}{c}{\texttt{\textbf{UNDERSPECIFIED}}} \\
\midrule
\textbf{Model} & 
{\textbf{Omniscient}} & 
{\textbf{Participant Centric}} & 
{\textbf{Overall}} & 
{\textbf{Omniscient}} \\
\midrule
\texttt{claude-3-5-sonnet-20241022} & $78.8$ & $50.5$ & $55.6$ & $63.8$ \\
\texttt{gpt-4o-2024-08-06} & $65.4$ & $34.8$ & $40.3$ & $54.3$ \\
\texttt{llama3-1-405b-instruct} & $67.0$ & $35.5$ & $41.2$ & $73.9$ \\
\texttt{llama3-2-90b-instruct} & $44.4$ & $21.7$ & $25.7$ & $44.4$ \\
\texttt{llama3-2-11b-instruct} & $9.1$ & $8.1$ & $8.3$ & $14.6$ \\
\bottomrule
\end{tabular}
\caption{\small{Model performance on (i) conversations with distractor (ii) underspecified conversations with unanswerable questions. For distractor conversations we compute accuracy for Omniscient, participant-centric and over all questions. For unswerable questions we only compute $\%$ unanswerability correctly identified. We compute this for only Omniscient questions as explained in section~\ref{subsec: unans_qu}}}
\label{tab:distractor_perf}
\vspace{-10pt}
\end{table*}
Our analysis of conversations with distractors (Table~\ref{tab:distractor_perf}) reveals a consistent drop in model performance across all models, especially for participant-centric type questions. 
For omniscient questions, the impact is relatively modest. \texttt{Claude-3.5} maintains strong performance with only $1.2\%$ drop in performance compared to that on base conversation. Similarly, \texttt{Llama3-405B} also shows only $2.2\%$ drop in performance. \texttt{GPT-4o} experiences a more noticeable performance drop of $\sim8\%$ to $65.4\%$ due to distractors.
The challenge becomes more pronounced with participant-centric questions, where models must simultaneously filter distractors while maintaining participant-specific information models. \texttt{Claude-3.5} achieves $50.5\%$ accuracy on these more complex scenarios, while \texttt{GPT-4o} and \texttt{Llama3-405B} attain $34.8\%$ and $35.5\%$ respectively.
These results highlight a critical limitation that current language models struggle to effectively filter out irrelevant information, especially when required to maintain a specific participant centric view of the conversation.
\vspace{-5pt}
\subsection{Handling Unanswerable Situations}\label{subsec: unans_qu}
\vspace{-5pt}
For underspecified conversations, we calculate the percentage of questions where models correctly identified the unanswerability. It's important to note that for underspecified conversations, if the participant-centric question is unanswerable, its omniscient variant will also be unanswerable. Therefore, a model could potentially arrive at the correct answer (identifying unanswerability) for a participant-centric question by using omniscient reasoning. So we only report performance on Omniscient questions.
As shown in Table~\ref{tab:distractor_perf}, models demonstrate varying abilities to recognize information gaps. \texttt{Claude-3.5} correctly identifies $63.8\%$ of unanswerable questions, while \texttt{Llama3-405B} performs surprisingly well at $73.9\%$. GPT-4o achieves $54.3\%$ accuracy, with smaller models showing significantly reduced capabilities. This demonstrates that even advanced models struggle to consistently recognize when they lack sufficient information to answer a question and are biased to generate answer even when it is not possible to do so.

\vspace{-10pt}
\section{Conclusion}
\vspace{-5pt}
We present \numtom{}, a novel multiparty conversation comprehension benchmark that tests for a subset of Theory of Mind (ToM) capabilities.
Our benchmark specifically targets tracking dynamic information states, managing knowledge asymmetries across multiple participants, and integrating this understanding with numerical reasoning—challenges that mirror real-world social interactions. Our evaluation of state-of-the-art language models reveals significant gaps in these types of ToM capabilities, where even strong models like Claude-3.5 struggle with participant-centric reasoning, and fare even worse in false belief settings.
Additionally, \numtom{} introduces a scalable approach for constructing complex, information-rich conversations with controlled information asymmetries. By leveraging a multi-stage generation process that combines LLM capabilities with structured templates sampled from a Markov process, we create conversations that systematically test specific aspects of ToM reasoning while maintaining coherence and realism. This methodology offers a blueprint for developing increasingly sophisticated conversation-based evaluation resources. These findings highlight a fundamental challenge: current language models struggle to maintain accurate mental models of different participants' knowledge states throughout extended conversations.
\numtom{} thus provides valuable insights into the current limitations of language models while establishing clear directions for improving ToM capabilities essential for effective human-AI interaction in multi-party settings.


\section{Acknowledgement}
This work is supported in part by DARPA for the KAIROS program under agreement number FA8750-19-2-1003 and by an Amazon Research Award. We also thank Prof. Katrin E. Erk, Prof. H. Andrew Schwartz, Prof. Jordan Kodner, and Prof. Owen Rambow for providing insightful feedback that was essential for improving this work.

\bibliography{colm2025_conference}
\bibliographystyle{colm2025_conference}

\appendix
\newpage
\section{Appendix}
\label{sec:appendix}

\subsection{Script Premise Components}\label{subsec:script2conv_ex}
\begin{figure}[h!]
    \centering
    \includegraphics[width=0.7\textwidth]{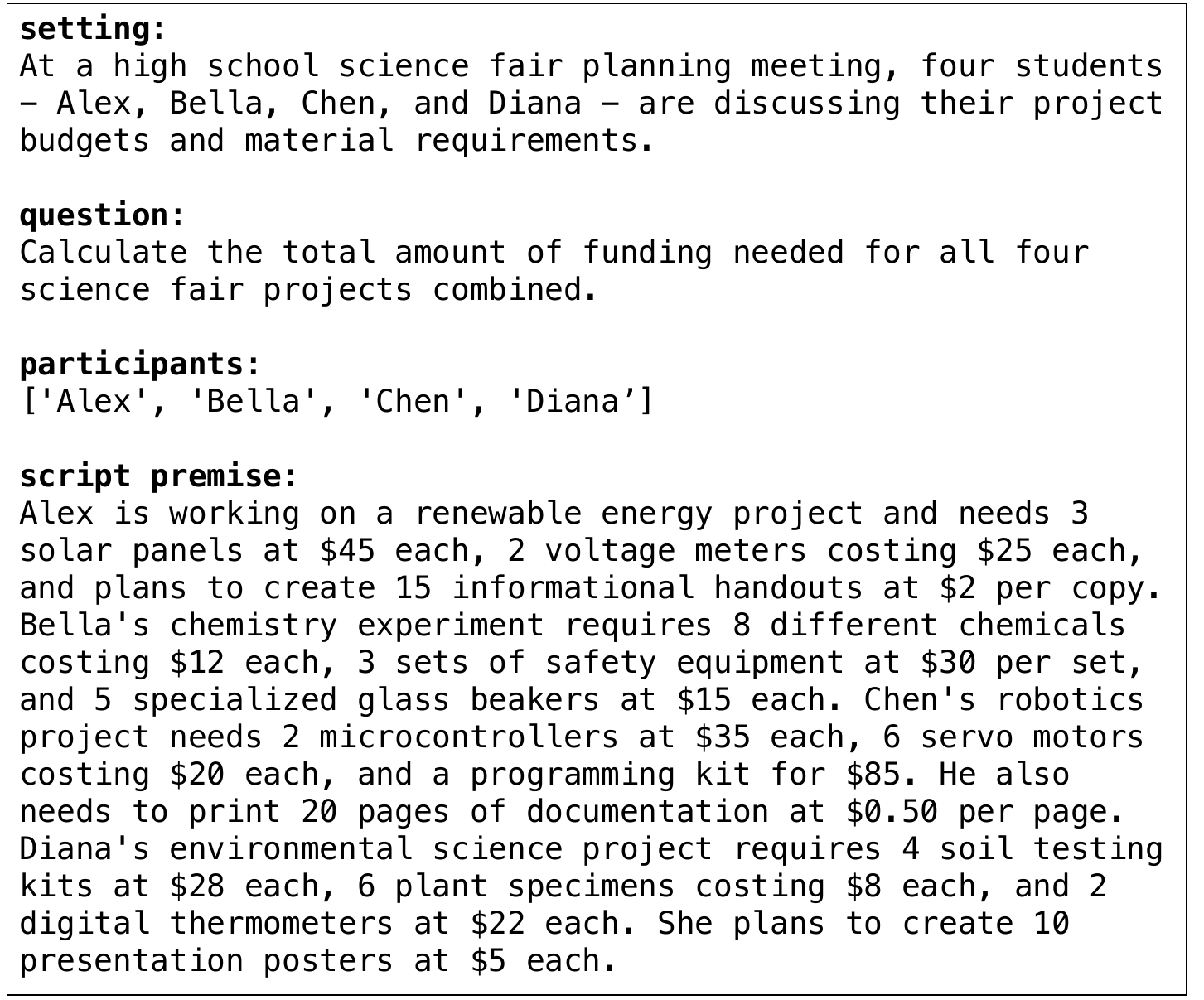}
    \caption{\textbf{Script Premise} Components for a Travel Agency Meeting with $4$ participants.}
    \label{Fig:premise_dict}
\end{figure}

\subsection{Variable State Perturbation}
\begin{figure}[h!]
    \centering
    \includegraphics[width=0.7\textwidth]{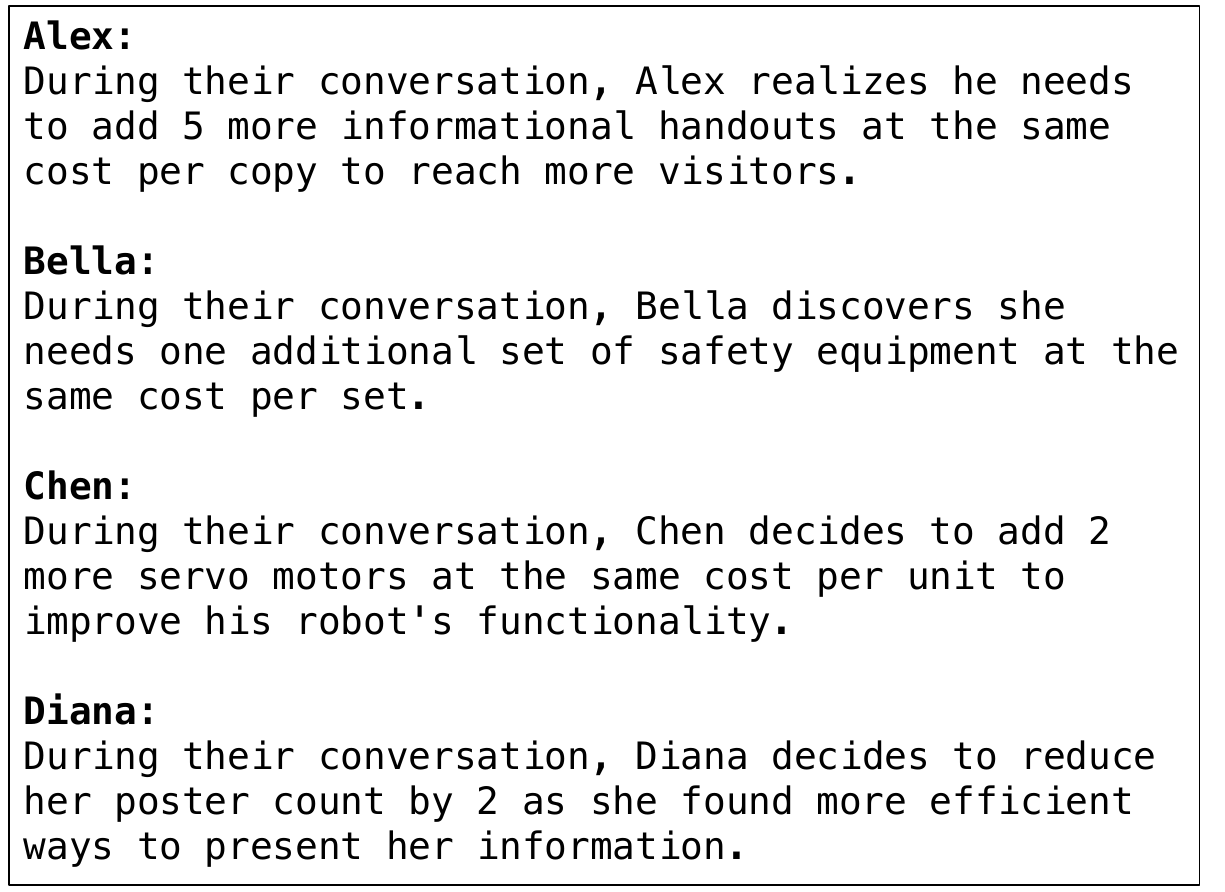}
    \caption{\textbf{Variable State Perturbation} information for each participant, that perturbs a variable state established in the script premise (Figure~\ref{Fig:premise_dict})}
    \label{Fig:var_change}
\end{figure}

\newpage
\subsection{Template Generation}
\begin{figure}[h!]
    \centering
    \includegraphics[width=0.6\textwidth]{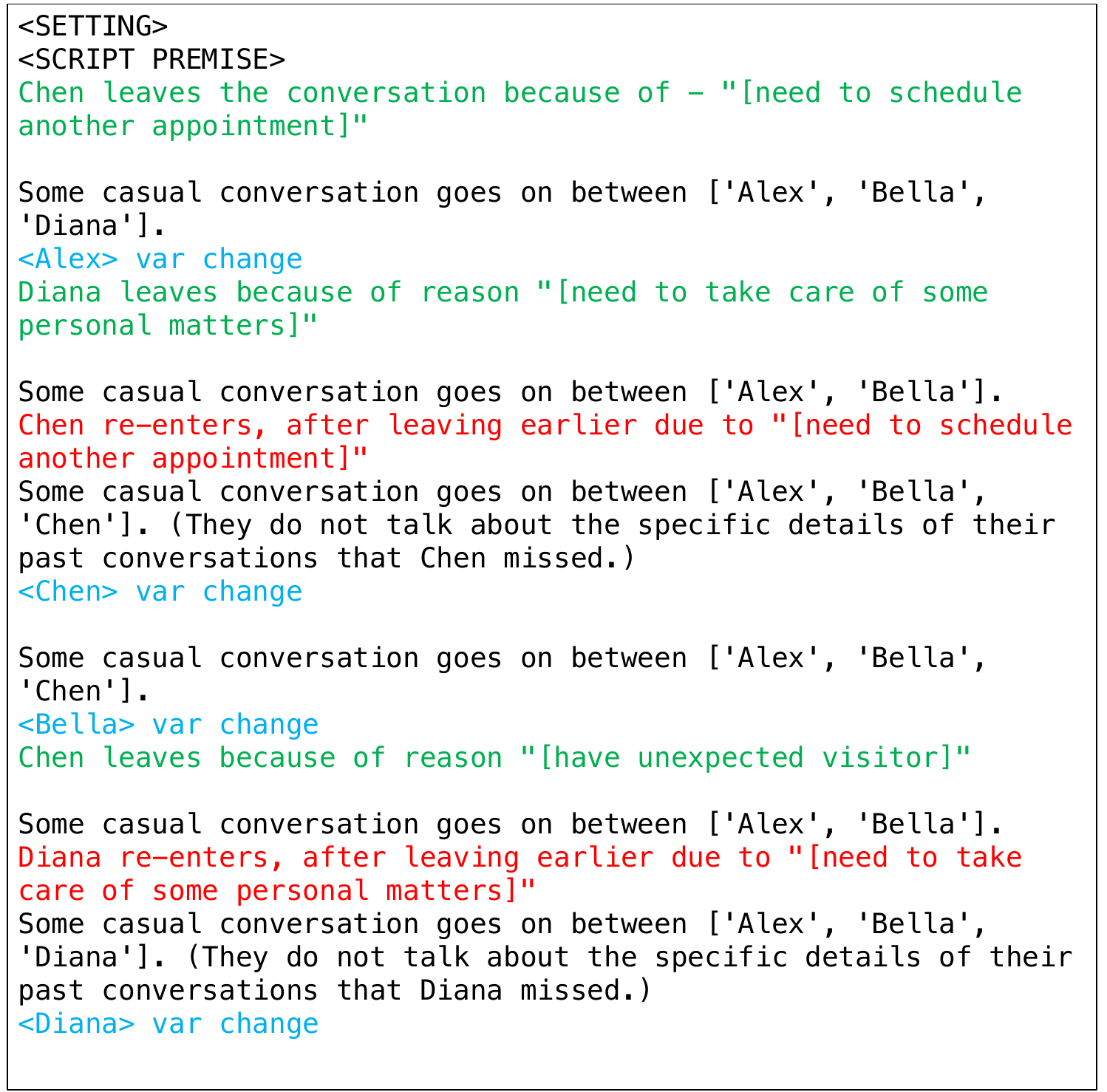}
    \caption{\textbf{Script Template} with participant movement (\textcolor{red}{red} and \textcolor{green}{green} colored) and slots (\textcolor{cyan}{\texttt{<name> var change}}) for pariable state perturbation information to be exchanged during the conversation.}
    \label{Fig:conv_template}
\end{figure}

\subsection{Assembling the Script}
\begin{figure}[h!]
    \centering
    \includegraphics[width=0.7\textwidth]{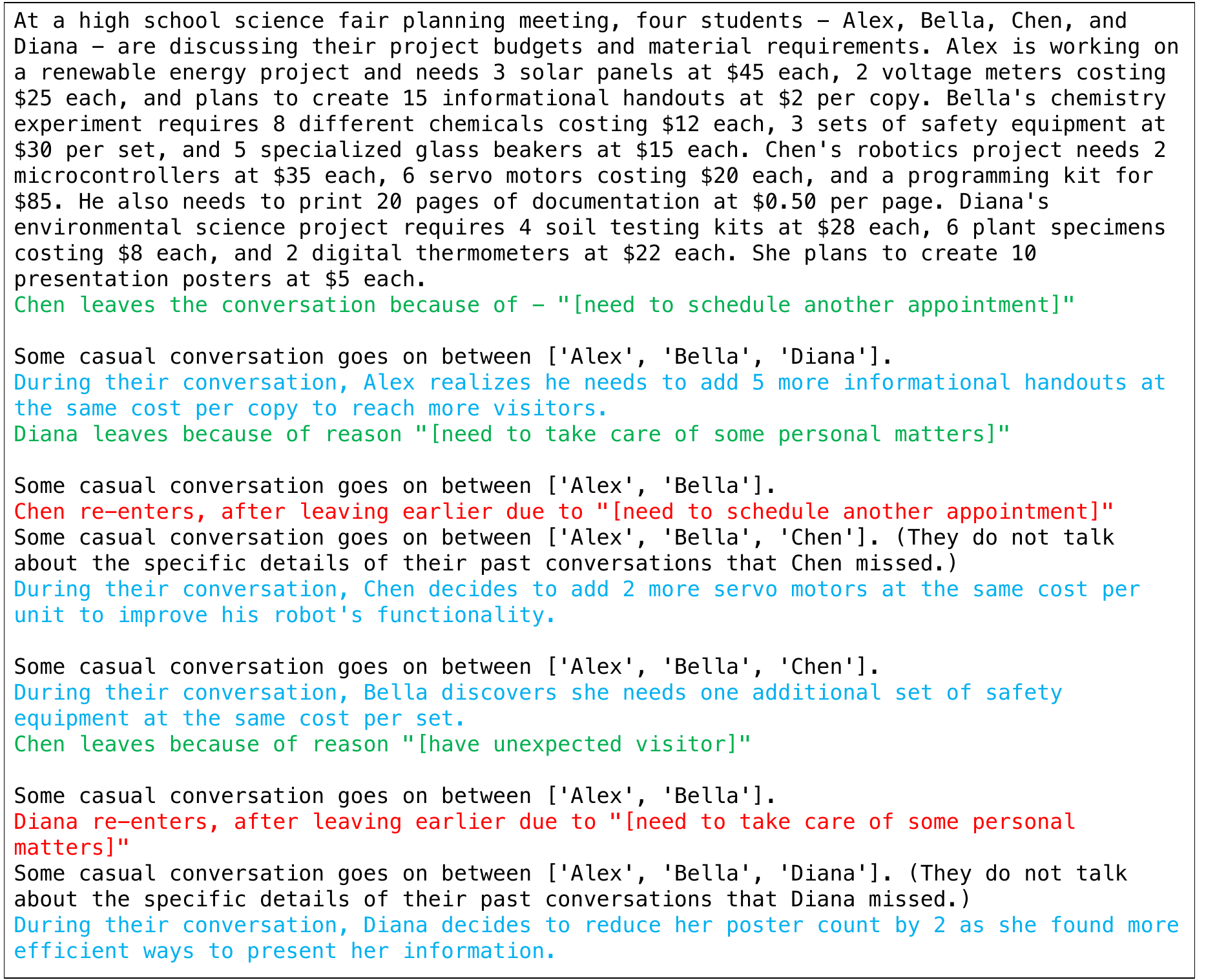}
    \caption{\textbf{Assmebling the script} by filling in the template slot (Figure~\ref{Fig:conv_template}) with relevant (i) setting, (ii) premise (Figure~\ref{Fig:premise_dict} and (iii) variable state perturbations (Figure~\ref{Fig:var_change}).}
    \label{Fig:full_script}
\end{figure}

\newpage
\subsection{Conversation Generation from Script}
\begin{figure}[h!]
    \centering
    \includegraphics[width=\textwidth]{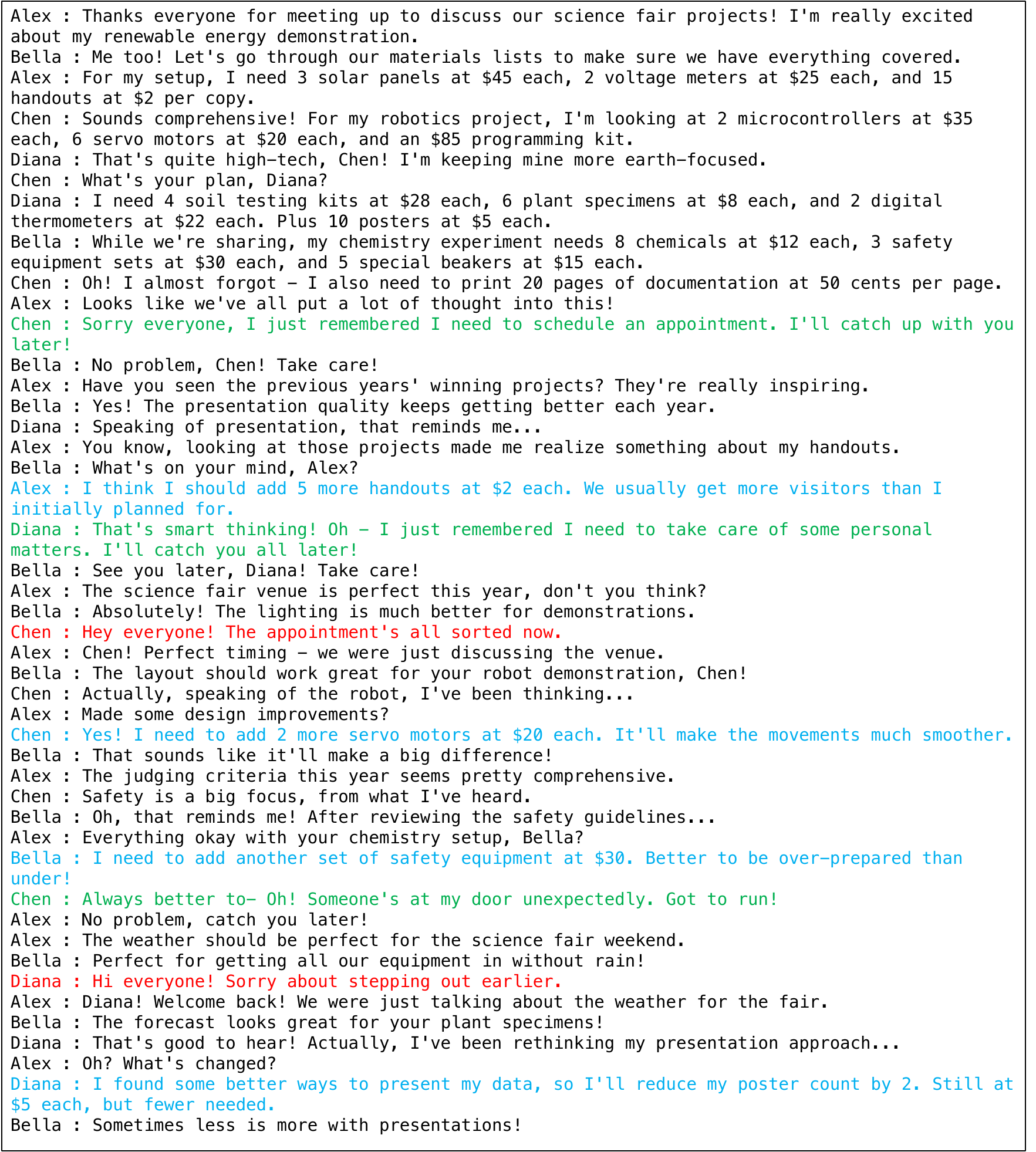}
    \caption{\textbf{Conversation from Script:} Final conversation based on the conversation script (Figure~\ref{Fig:premise_dict}).}
    \label{Fig:conversation_c11}
\end{figure}

\subsection{Script Template Generation}\label{appx:script_tmplt_grmmr}
\begin{figure}[ht!]
    \centering
    \includegraphics[width=0.8\textwidth]{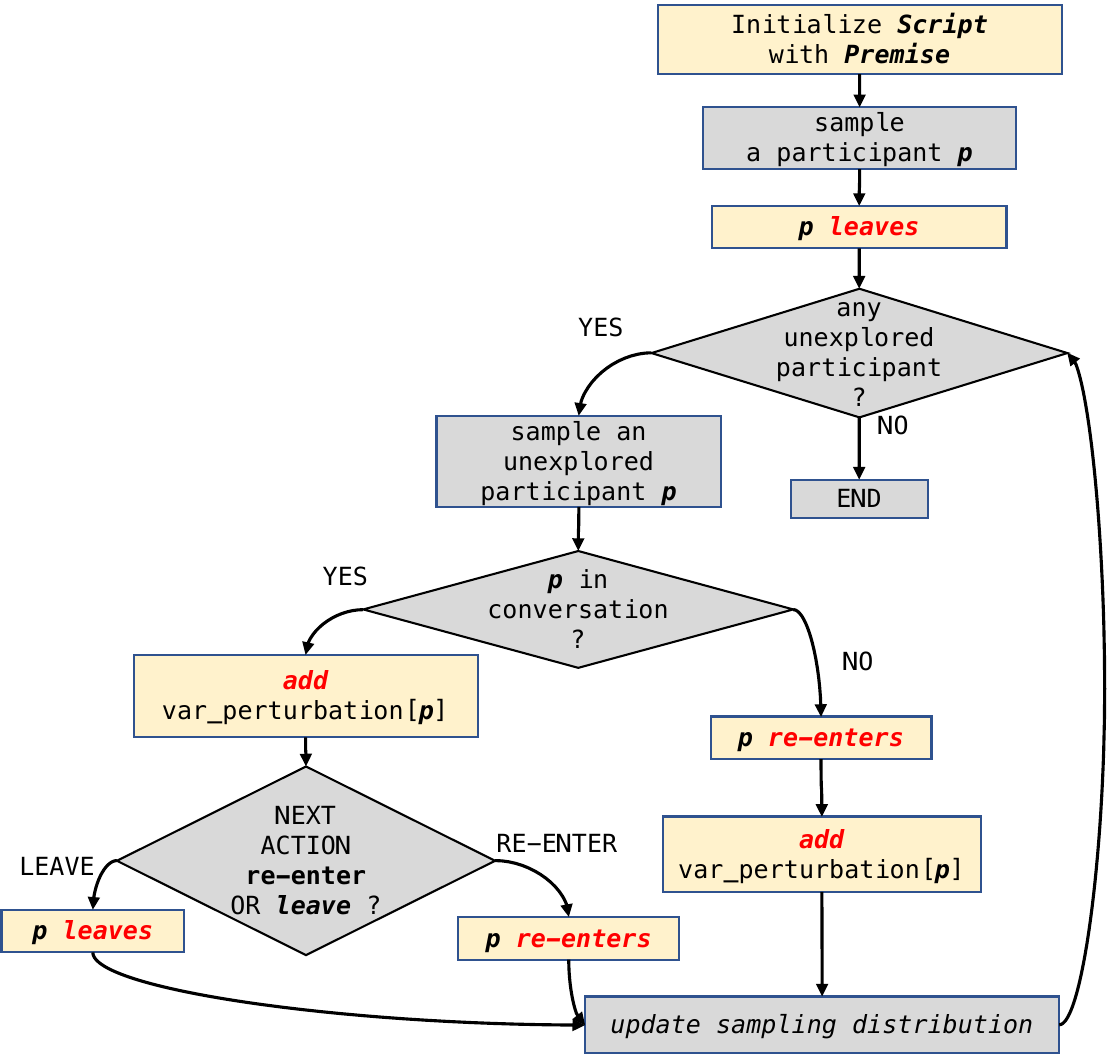}
\caption{Markov Chain process for generating the discourse of the script which will be translated into a conversation for the data in~\numtom{}.}
    \label{Fig:script_tmplt_grmmr}
\end{figure}
As shown in Figure~\ref{Fig:script_tmplt_grmmr}, the generation follows a Markov Chain process that creates dynamic script templates based on the number of participants, and a set of leaving reasoning mentioned in \cite{kim-etal-2023-fantom}. 
The process begins by establishing a premise slot that sets the foundation for the conversation. Initially, all participants have an equal probability of leaving the conversation.
The generation unfolds in several steps:
\begin{enumerate}
    \item First, a participant is randomly selected to leave the conversation, with their departure reason chosen from a predefined list of plausible excuses.
    \item Next, the generator selects a participant to share information (variable state perturbation) from those who haven't yet contributed to the conversation. This can happen in two ways:
    \begin{itemize}
        \item If the selected participant is currently present in the conversation, they directly share their information
        \item If the selected participant is absent, they first re-enter the conversation before sharing their information
    \end{itemize}
    \item After each information sharing event, the generator randomly decides between two possible actions:
    \begin{itemize}
        \item Having another participant leave the conversation
        \item Having a previously departed participant re-enter the conversation
    \end{itemize}
\end{enumerate}

To make the conversation flow more natural, the probability of a participant leaving again is reduced by $0.75$ each time they exit the conversation. This prevents unrealistic patterns of repeated exits and entries by the same participant. Between these major events (leaving, re-entering, and information sharing), the generator includes segments of casual conversation among the current participants. When a participant re-enters, the generator explicitly notes that the others don't discuss the specific details of conversations that occurred during their absence.

\newpage
\subsection{Benchmarking Models on their Complete Set of Data with Valid Prediction}\label{appx:full_result}
\renewcommand{\arraystretch}{1.1}
\begin{table*}[!h]
\centering
\begin{tabularx}{\textwidth}{l p{0.16\textwidth} p{0.16\textwidth} p{0.16\textwidth}}
\toprule
\multicolumn{4}{c}{\texttt{MODEL PERFORMANCE ON BASE CONVERSATIONS (FULL DATA)}} \\
\toprule
\textbf{Model} & 
{\textbf{Omniscient \newline (\%)}} & 
{\textbf{Participant \newline Centric (\%)}} & 
{\textbf{Overall \newline (\%)}}\\
\midrule
\texttt{claude-3-5-sonnet-20241022} & $82.6$ & $57.4$ & $81.8$ \\
\texttt{gpt-4o-2024-08-06} & $69.4$ & $40.9$ & $45.95$ \\
\texttt{llama3-1-405b-instruct} & $66.0$ & $36.0$ & $41.2$ \\
\texttt{llama3-2-90b-instruct} & $48.5$ & $23.7$ & $27.94$ \\
\texttt{llama3-2-11b-instruct} & $9.8$ & $8.7$ & $8.9$ \\
\bottomrule
\end{tabularx}
\caption{Model accuracy ($\%$) on regular/base conversations without any distractor. Numbers reported are for models' performance on their full set of valid, parsable predictions.}
\label{tab:base_perf_full}
\end{table*}
\renewcommand{\arraystretch}{1.1}
\begin{table*}[h!]
\small
\centering
\begin{tabular}{@{} l | ccc | c @{}}
\toprule
\multicolumn{5}{c}{\texttt{MODEL ACCURACY (\%) ON CONVERSATIONS (FULL DATA)}} \\
\midrule
& \multicolumn{3}{c|}{\texttt{\textbf{DISTRACTOR}}} & 
\multicolumn{1}{c}{\texttt{\textbf{UNDERSPECIFIED}}} \\
\midrule
\textbf{Model} & 
{\textbf{Omniscient}} & 
{\textbf{Participant Centric}} & 
{\textbf{Overall}} & 
{\textbf{Omniscient}} \\
\midrule
\texttt{claude-3-5-sonnet-20241022} & $78.1$ & $55.7$ & $24.$ & $58.1$ \\
\texttt{gpt-4o-2024-08-06} & $63.7$ & $35.6$ & $40.4$ & $44.6$ \\
\texttt{llama3-1-405b-instruct} & $65.0$ & $35.3$ & $40.6$ & $70.3$ \\
\texttt{llama3-2-90b-instruct} & $43.8$ & $22.1$ & $25.8$ & $44.3$ \\
\texttt{llama3-2-11b-instruct} & $9.1$ & $8.5$ & $8.6$ & $13.0$ \\
\bottomrule
\end{tabular}
\caption{Model accuracy (\%) on conversations with distractor and underspecified conversations where questions are unanswerable. The distractor conversations we compute accuracy for Omniscient, participant-centric and over all questions. For unswerable questions we only compute $\%$ unanswerable }
\label{tab:distractor_perf_full}
\end{table*}
We report models performance in Table~\ref{tab:base_conv_perf}, \ref{tab:distractor_perf} on a common subset of data where all models provided valid responses. Here we report the models' performance based on their full set of valid, parsable predictions across the entire dataset. Table~\ref{tab:base_perf_full}, \ref{tab:distractor_perf_full} reports the model performance on regular/base, distractor and unanswerable conversation in \numtom{}.

\subsection{Dataset Generation}\label{subsec:dataset_genr}
Our complete dataset generation pipeline uses LLMs in a multi-stage fashion to generate the conversation, question, answer ($C, q, a$). Given the mathematically nuanced and constrained nature of our dataset, we use the latest state of art models \texttt{claude-3-5 sonnet 20241022}, and \texttt{gpt-4o 2024-08-06} with few-shot prompting for generating the required components of the dataset. For translating the script into their corresponding conversation, we use \texttt{claude-3-5-sonnet-20241022} with 1-shot prompting as we observed it to generate more more natural reading and longer conversations compared to \texttt{gpt-4o-2024-08-06}. However, to show that the relative performance trend of a model is not result of choice of model for translating script to conversation, we also generate data that uses \texttt{gpt-4o-2024-08-06} as script to conversation translation model and report result on it in the Appendix~\ref{appx:gpt4o_result}.

\newpage
\subsection{Benchmarking on Data with \texttt{gpt-4o} as Script to Conversation Translator }~\label{appx:gpt4o_result}
\renewcommand{\arraystretch}{1.1}
\begin{table*}[h!]
\centering
\begin{tabularx}{\textwidth}{l p{0.16\textwidth} p{0.16\textwidth} p{0.16\textwidth}}
\toprule
\multicolumn{4}{c}{\texttt{MODEL PERFORMANCE ON BASE CONVERSATIONS}}\\
\multicolumn{4}{c}{\texttt{GPT-4o TRANSLATED CONVERSATION}}\\
\toprule
\textbf{Model} & 
{\textbf{Omniscient \newline (\%)}} & 
{\textbf{Participant \newline Centric (\%)}} & 
{\textbf{Overall \newline (\%)}}\\
\midrule
\texttt{claude-3-5-sonnet-20241022} & $66.7$ & $43.0$ & $47.1$ \\
\texttt{gpt-4o-2024-08-06} & $57.1$ & $38.0$ & $41.3$ \\
\texttt{llama3-1-405b-instruct} & $61.9$ & $33.0$ & $38.0$ \\
\texttt{llama3-2-90b-instruct} & $47.6$ & $27.0$ & $30.6$ \\
\texttt{llama3-2-11b-instruct} & $14.3$ & $14.0$ & $14.0$ \\
\bottomrule
\end{tabularx}
\caption{Model accuracy ($\%$) on regular/base conversations without any distractor. Numbers reported here are for a common subset of conversations that are translated from script using \texttt{gpt-4o}.}
\label{tab:base_perf_gpt}
\end{table*}
\renewcommand{\arraystretch}{1.1}
\begin{table*}[h!]
\centering
\begin{tabularx}{\textwidth}{l p{0.16\textwidth} p{0.16\textwidth} p{0.16\textwidth}}
\toprule
\multicolumn{4}{c}{\texttt{MODEL PERFORMANCE ON DISTRACTOR CONVERSATIONS}}\\
\multicolumn{4}{c}{\texttt{GPT-4o TRANSLATED CONVERSATION}}\\
\toprule
\textbf{Model} & 
{\textbf{Omniscient \newline (\%)}} & 
{\textbf{Participant \newline Centric (\%)}} & 
{\textbf{Overall \newline (\%)}}\\
\midrule
\texttt{claude-3-5-sonnet-20241022} & $54.2$ & $34.3$ & $38.0$ \\
\texttt{gpt-4o-2024-08-06} & $41.7$ & $23.8$ & $27.1$ \\
\texttt{llama3-1-405b-instruct} & $41.7$ & $21.9$ & $25.6$ \\
\texttt{llama3-2-90b-instruct} & $20.8$ & $16.2$ & $17.1$ \\
\texttt{llama3-2-11b-instruct} & $8.3$ & $5.8$ & $6.3$ \\
\bottomrule
\end{tabularx}
\caption{Model accuracy ($\%$) on conversations with distractors. Numbers reported here are for a common subset of conversations that are translated from script using \texttt{gpt-4o}.}
\label{tab:distractor_perf_gpt}
\end{table*}
In this section, we report the model performance on question for \numtom{} conversations, where \texttt{gpt-4o} is used to translate the script to conversation. Table~\ref{tab:base_perf_gpt} reports the model performance on base/regular conversations, Table~\ref{tab:distractor_perf_gpt} on conversations with distractor and on underspecified conversations for which the questions are unanswerable. We do not include these data as part of \numtom{} dataset, due to the observed quality of the translated conversations to be erroneous and unnatural in generation. However, we are reporting the performance just to show that the relative performance trends of the model is same as that observed for conversations generated with \texttt{Claude-3.5} (Table~\ref{tab:base_conv_perf}, \ref{tab:distractor_perf}).

\newpage
\subsection{Prompts}\label{subsec:prompts}

\subsubsection{Script Premise Components}
\begin{figure*}[h!]
    \centering
    \includegraphics[width=\textwidth]{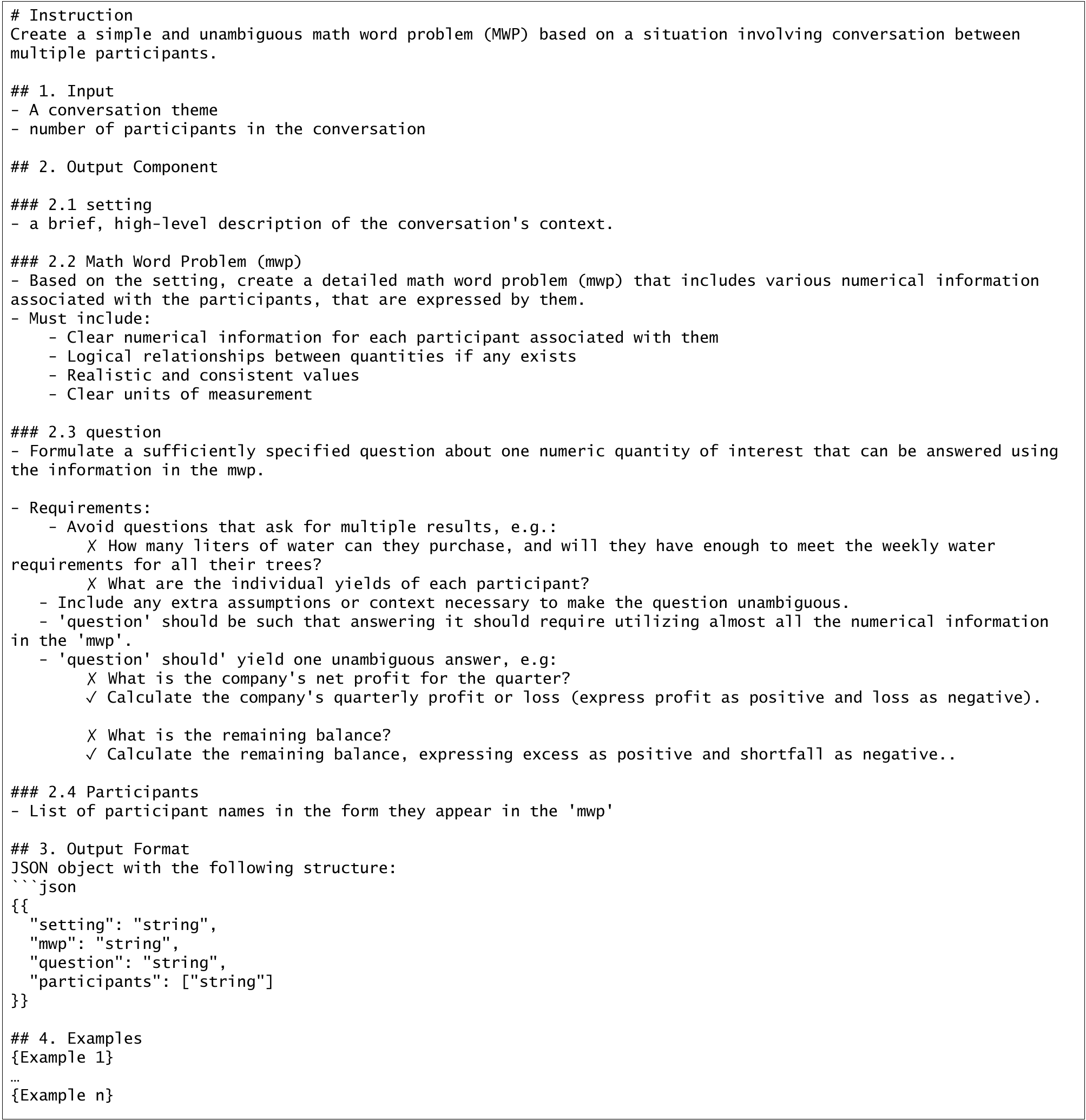}
\caption{Conversation script premise prompt}
    \label{Fig:premise_prompt}
\end{figure*}

\newpage
\subsubsection{Variable State Perturbations Generation}
\begin{figure*}[h!]
    \centering
    \includegraphics[width=1\textwidth]{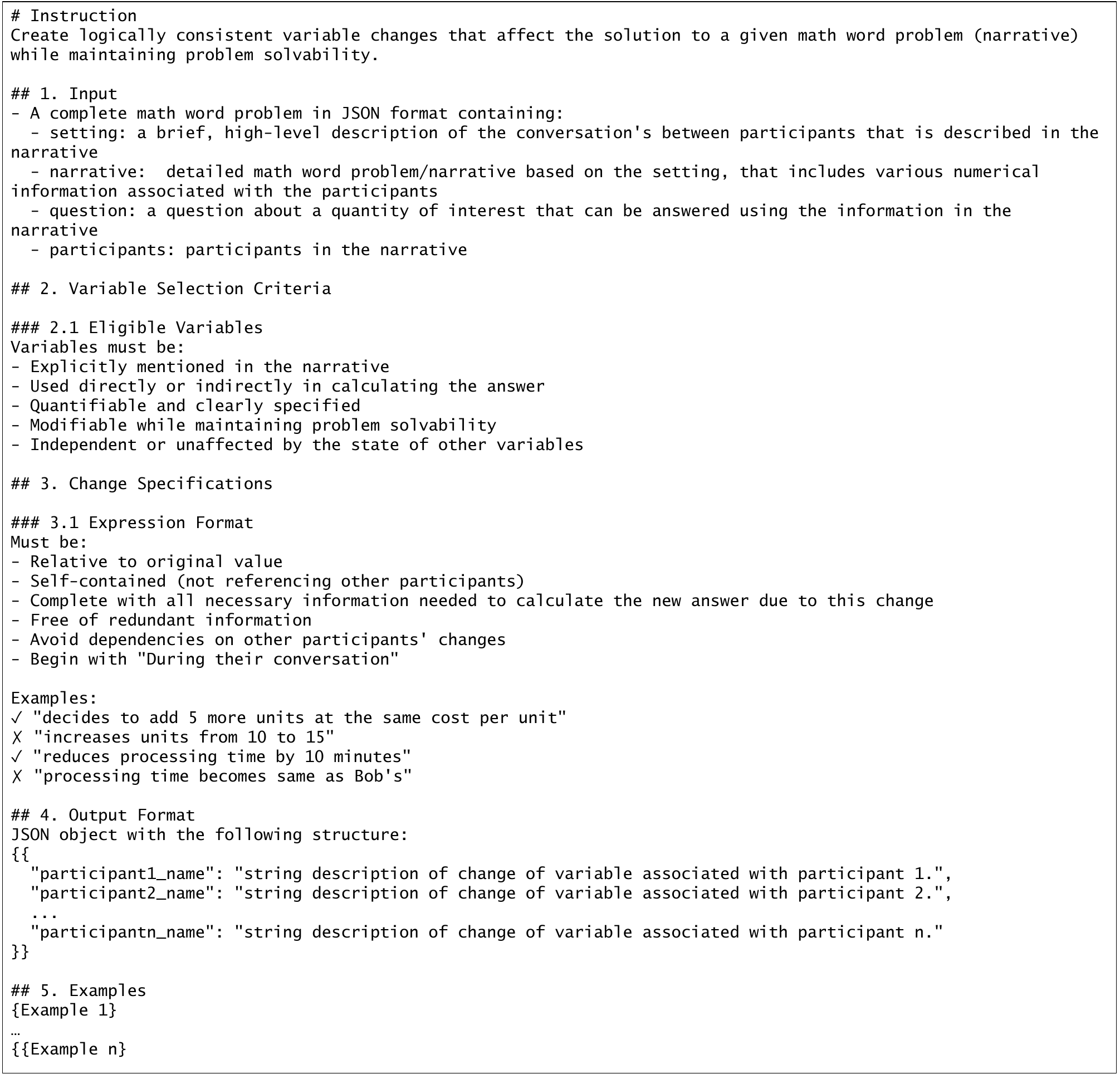}
\caption{Variable State Perturbation: Information perturbing the state of the variables established in the remise}
    \label{Fig:var_ch_prompt}
\end{figure*}

\newpage
\subsubsection{Generate Underspecified and Distractor Script Variants}\label{subsec:under_distr_descr}
\begin{figure*}[h!]
    \centering
    \includegraphics[width=1\textwidth]{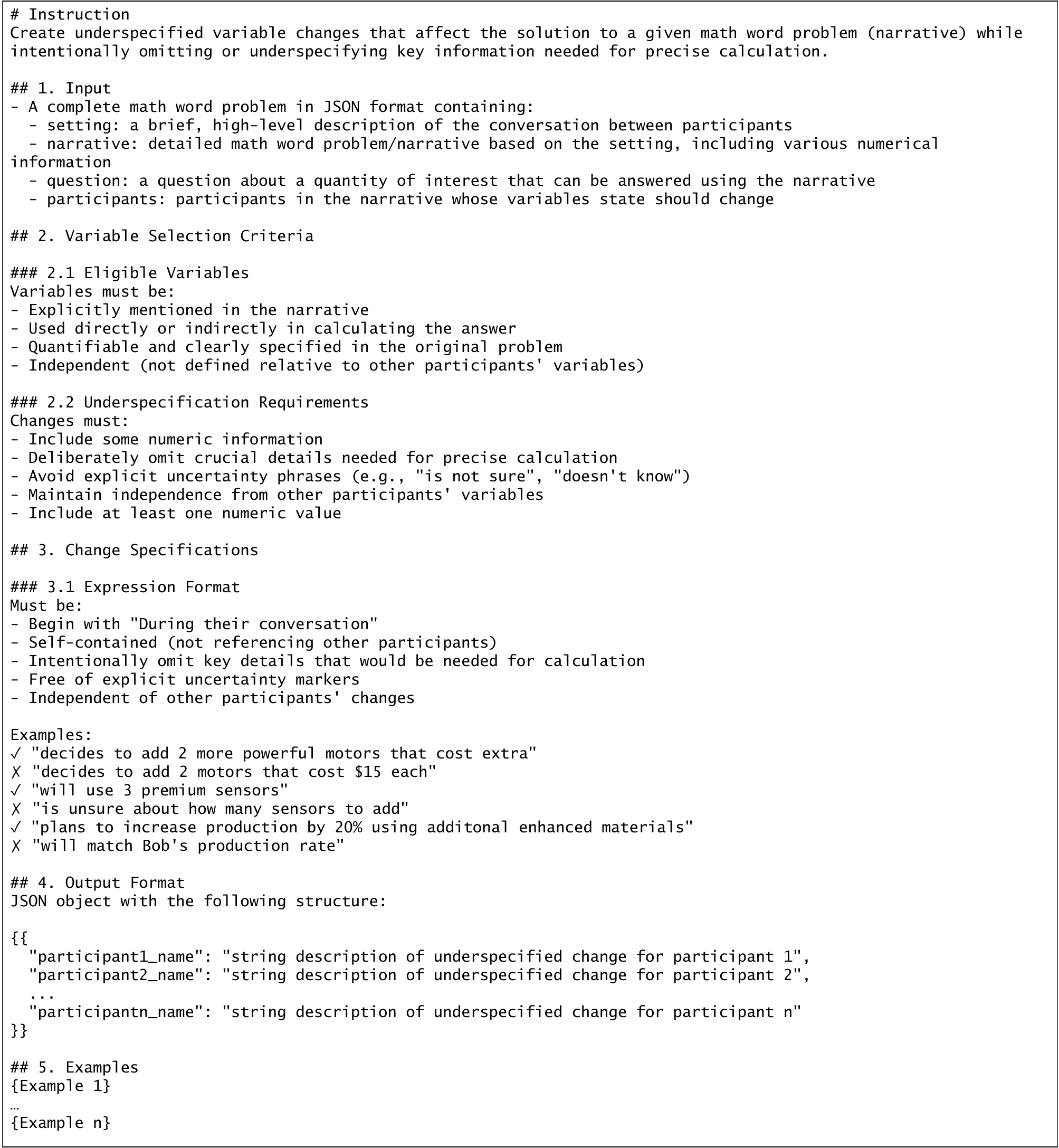}
\caption{Prompt for generating underspecified variable state perturbation information that effects the answer to the question}
    \label{Fig:underspec_var_ch_prompt}
\end{figure*}

\begin{figure}[h!]
    \centering
    \includegraphics[width=\textwidth]{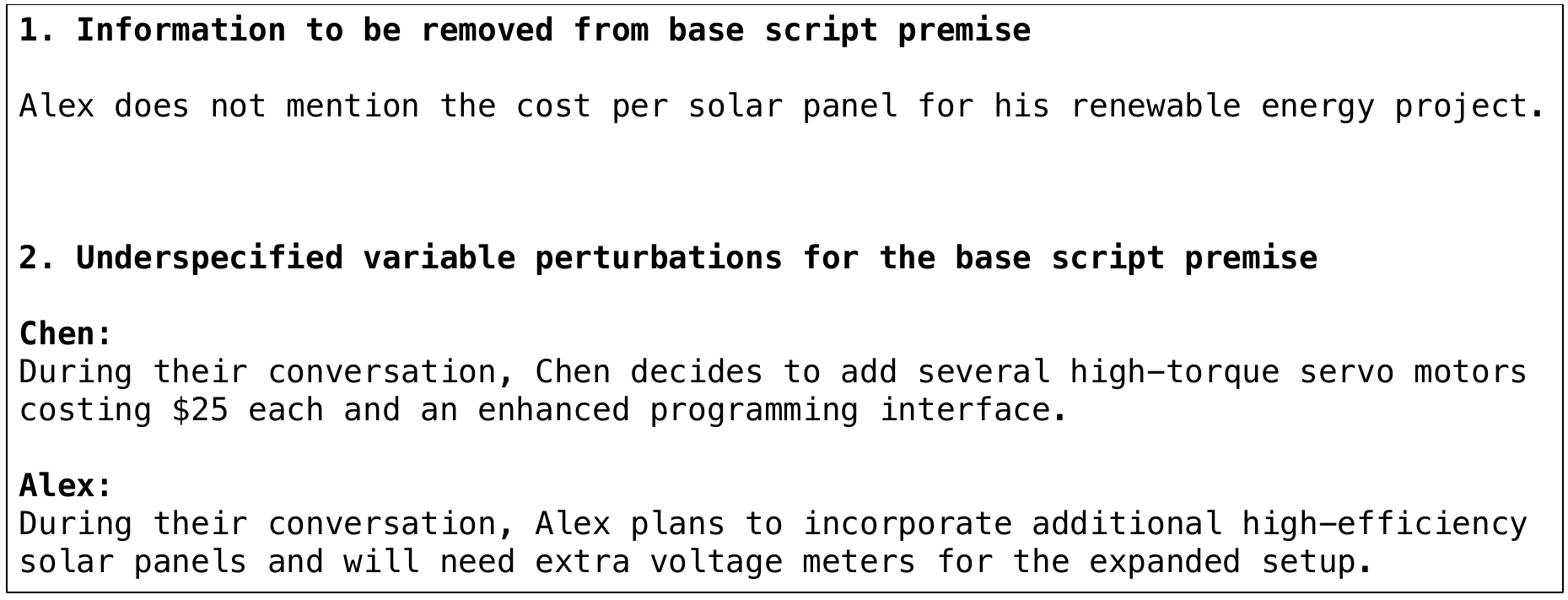}
    \caption{Components Required for Unanswerable Script Generation: (1) Elements to be excluded from the base script premise (Figure~\ref{Fig:premise_dict}) to create an underspecified premise, and (2) Underspecified variable state perturbations for the base premise.}
    \label{Fig:underspec_comps}
\end{figure}

\paragraph{Underspecific Script}
We generate Unanswerable or underspecified variants $S^-$ of base script $S$ through systematic information omission. Given a base script $S$ with premise $S_p$, we create an underspecified premise $S_p^-$ (Figure~\ref{Fig:underspec_comps} (1)) by removing key information required to answer the question. We then generate two sets of variable state perturbations: (i) $V_{S_p^-}$ for the underspecified premise $S_p^-$ and (ii) $V_{S_p}^-$ (Figure~\ref{Fig:underspec_comps} (2)) for the base script premise $S_p$ with underspecified variable information. We create unanswerable script $S^-$ by combining ($S_p^-$, $V_{S_p^-}$) or ($S_p$, $V_{S_p}^-$, $V_{S_p}$) through our script template, that test models' ability to recognize insufficient information. The components for the distractor scripts are generated with few-shot prompting using a Large Language Model. The prompts for generating $S^-_p$ and $V_{S_p}^-$ are shown in Appendix, Figure~\ref{Fig:distr_undersp_premise_prompt}, and \ref{Fig:underspec_var_ch_prompt}.  

\begin{figure}[h!]
    \centering
    \includegraphics[width=0.8\textwidth]{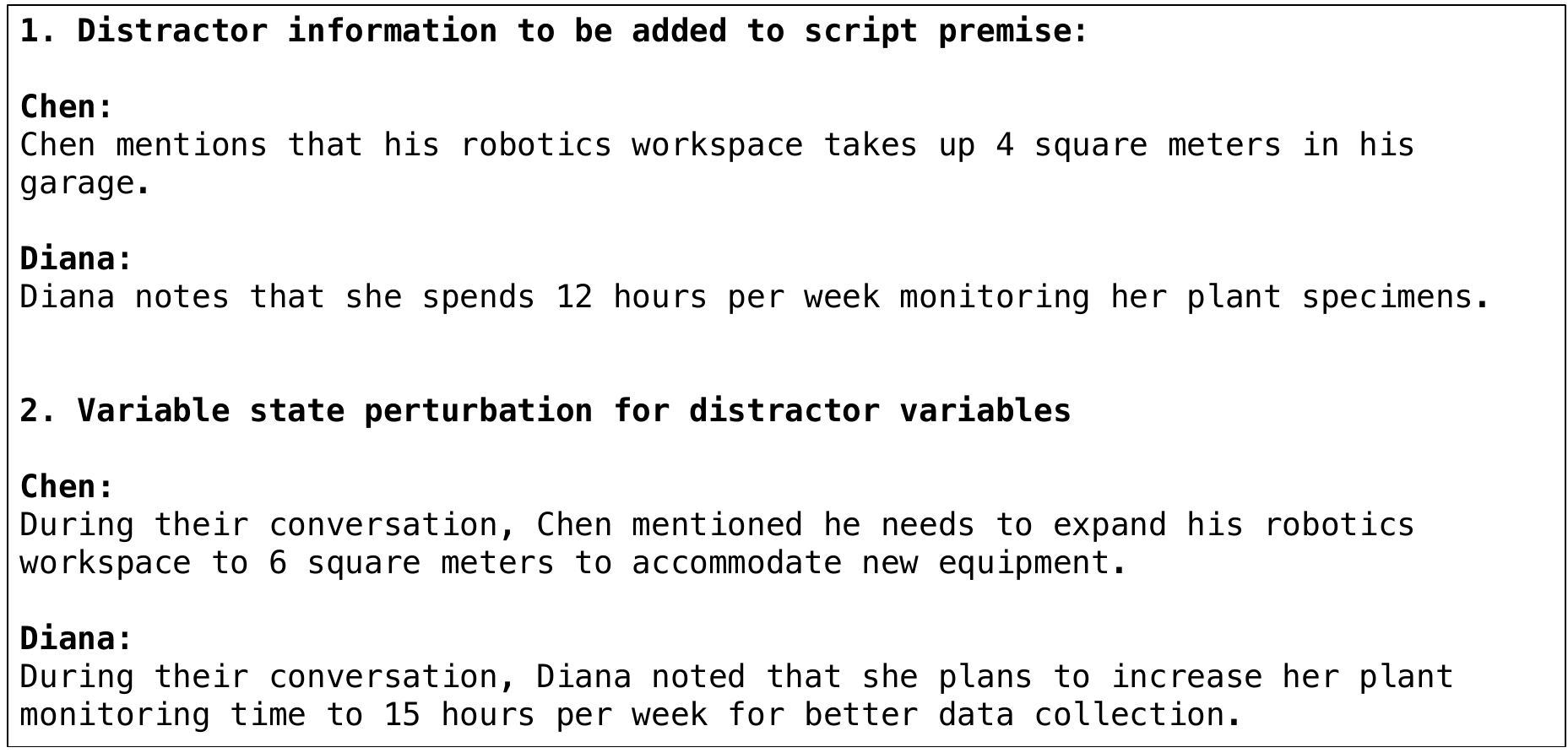}
    \caption{Components Required for Distractor Script Generation: (1) Distractor variable information to be added to the base script's premise (Figure~\ref{Fig:premise_dict}) to create a distractor premise, and (2) Variable state perturbations for the distractor variables in the distractor premise.}
    \label{Fig:distractor_comps}
\end{figure}

\paragraph{Distractor Script} For generating the distractor variants $S^+$, we augment the base script $S$ with additional, thematically relevant but question-irrelevant information. We first transform a base script premise $S_p$ into an augmented premise $S_p^+$ by introducing information about new quantities $u_i$ and $u_j$ (Figure~\ref{Fig:distractor_comps} (1)) associated with participants $p_i$ and $p_j$. These additions maintain conversational coherence while being irrelevant to answering the target question. We then generate $u_i$ and $u_j$'s  corresponding variable state perturbations $v_i^+$ and $v_j^+$ (Figure~\ref{Fig:distractor_comps} (2)) for these new quantities. The final distractor script $S^+$ is assembled using an augmented template that augments the original template slots with positions for distractor variable updates. This process preserves the original answer to question $q$, as the added information in $S^+$ is designed to be independent of the solution path. Similar to the distractor scripts, the components for the unanswerable scripts are generated with few-shot prompting using a Large Language Model. The prompt template used to generate $S_p^+$, and $v_i^+$ and $v_j^+$ is shown in Appendix, Figure~\ref{Fig:distr_undersp_premise_prompt}, \ref{Fig:distr_var_ch_prompt}.

\begin{figure*}[h!]
    \centering
    \includegraphics[width=1\textwidth]{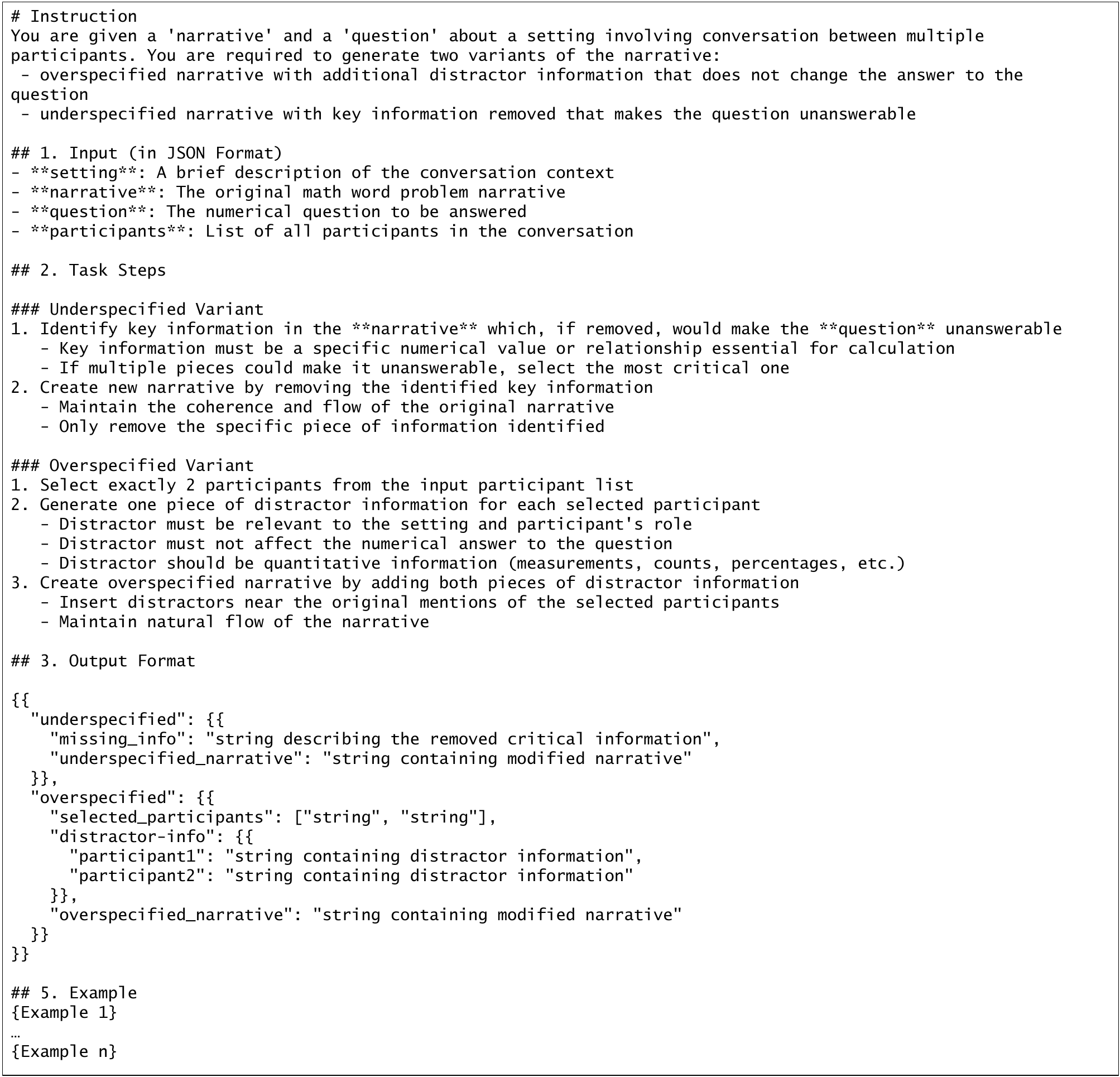}
\caption{Prompt for generating (i) underspecified premise and (ii) premise with distractor information.}
    \label{Fig:distr_undersp_premise_prompt}
\end{figure*}

\newpage
\subsubsection{Distractor Variable State Perturbations Generation}
\begin{figure*}[h!]
    \centering
    \includegraphics[width=1\textwidth]{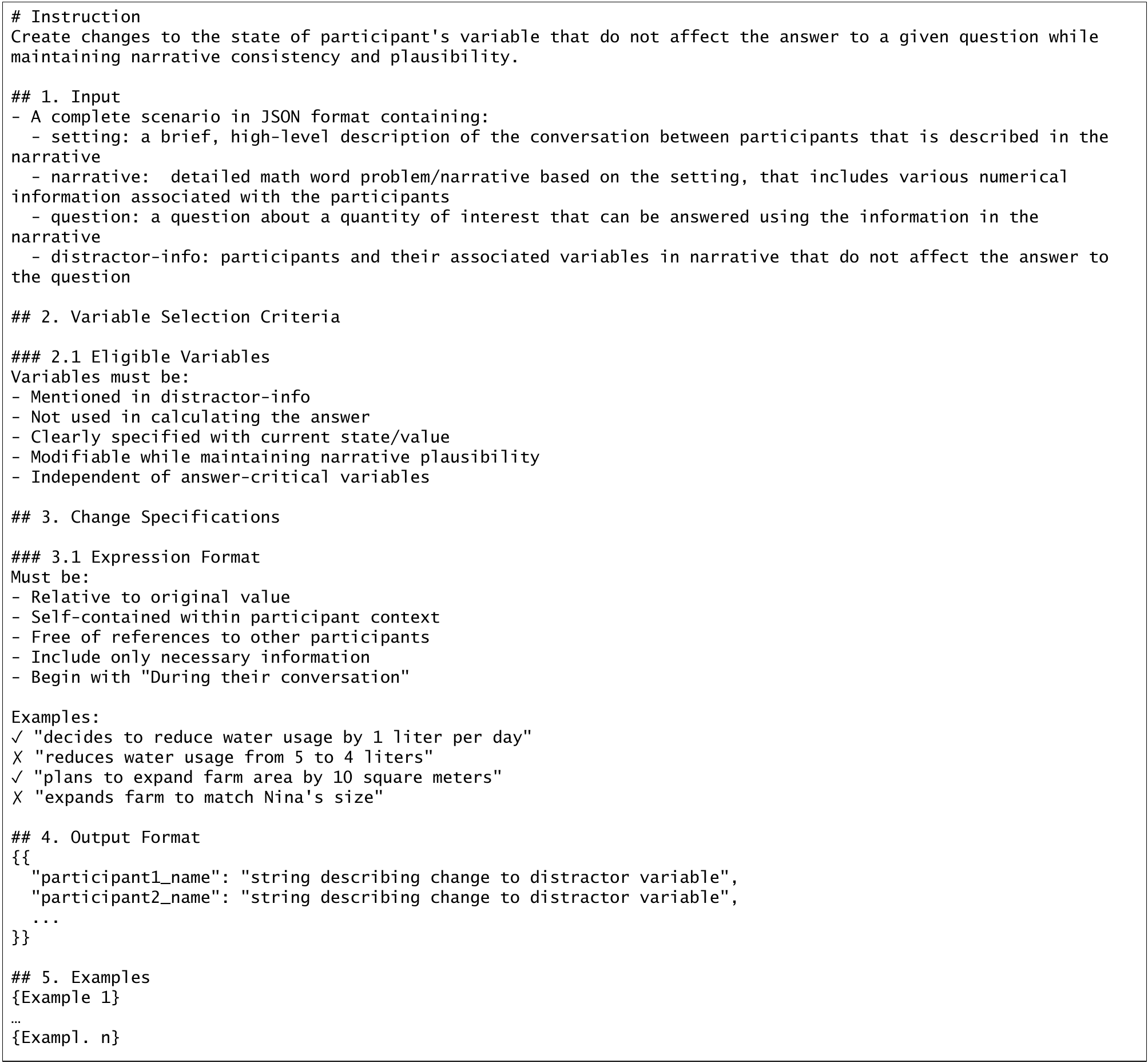}
\caption{Perturbing distractor information in premise with distractor}
    \label{Fig:distr_var_ch_prompt}
\end{figure*}

\newpage
\subsubsection{Underspecified Variable State Perturbations Generation}

\newpage
\subsubsection{Script to Conversation Translation}
\begin{figure*}[h!]
    \centering
    \includegraphics[width=1\textwidth]{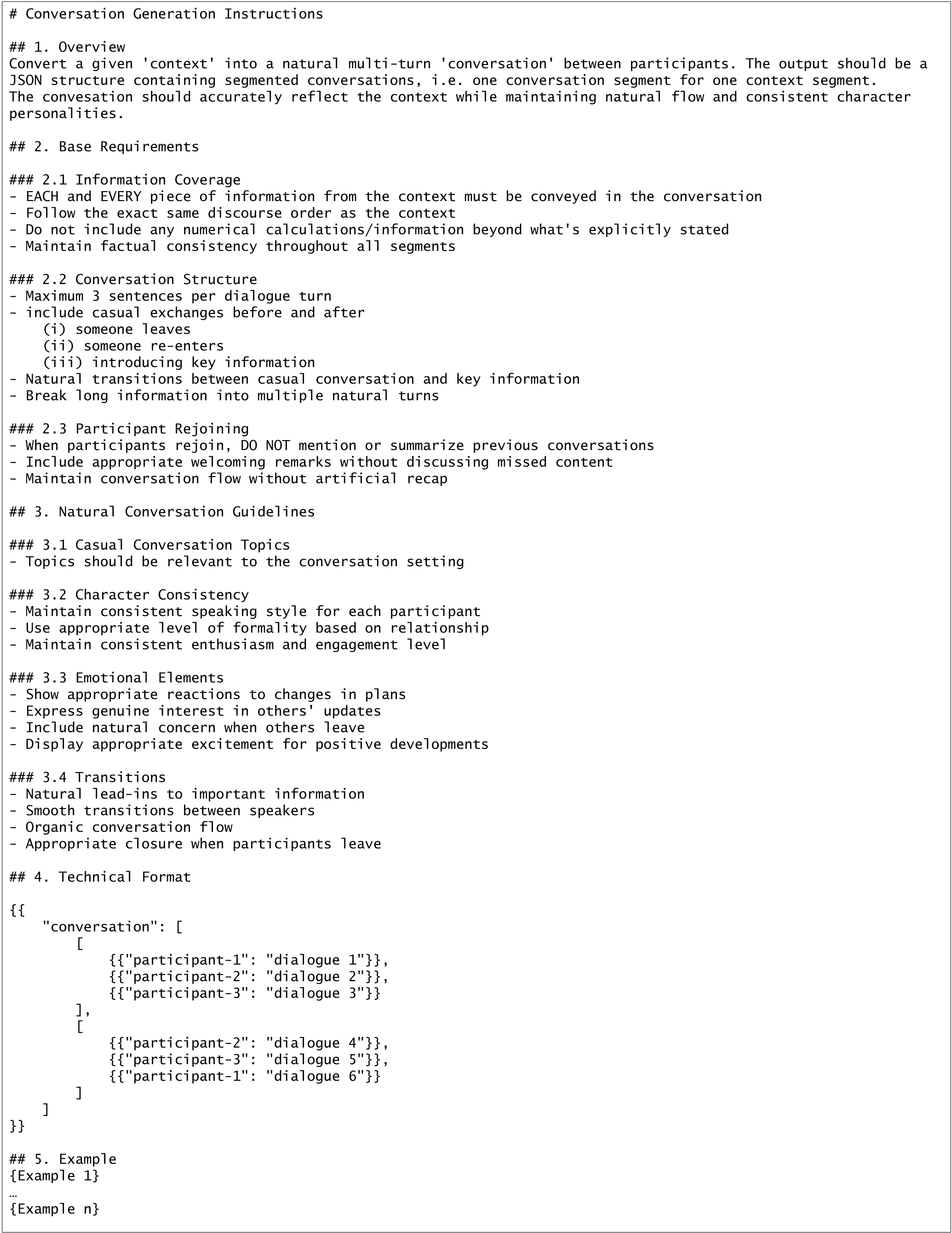}
\caption{Prompt for translating script to natural conversation.}
    \label{Fig:script2conv_prompt}
\end{figure*}

\newpage
\subsubsection{Model Inference}
\begin{figure*}[h!]
    \centering
    \includegraphics[width=\textwidth]{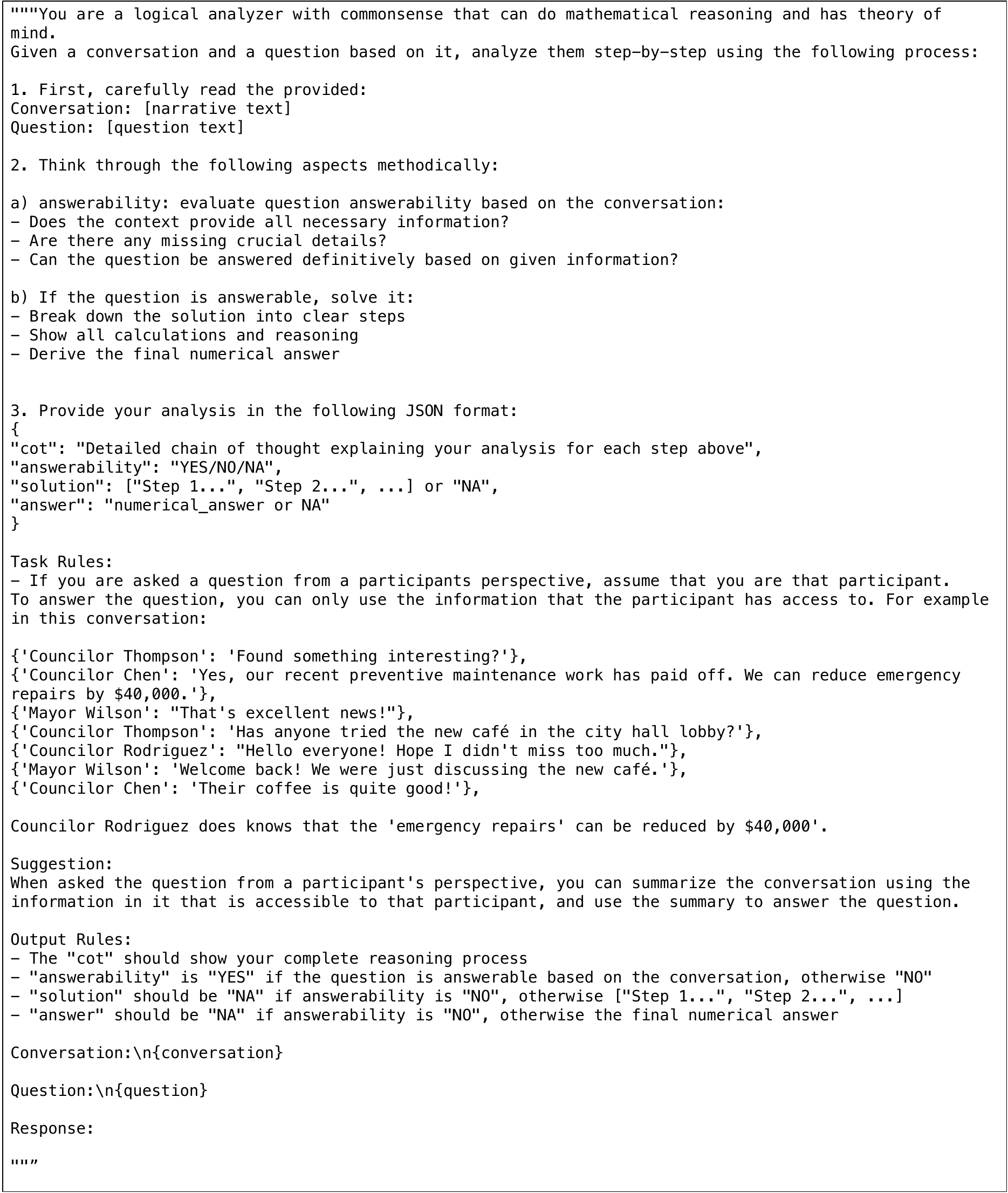}
\caption{Prompt used for performing inference for the task.}
    \label{Fig:inference_prompt}
\end{figure*}

\newpage
\subsection{Distractor and Unanswerable Conversations}\label{subsec:distr_upder_spec_ex}

\subsubsection{Distractor Conversations}
\begin{figure}[h!]
    \centering
    \includegraphics[width=0.8\textwidth]{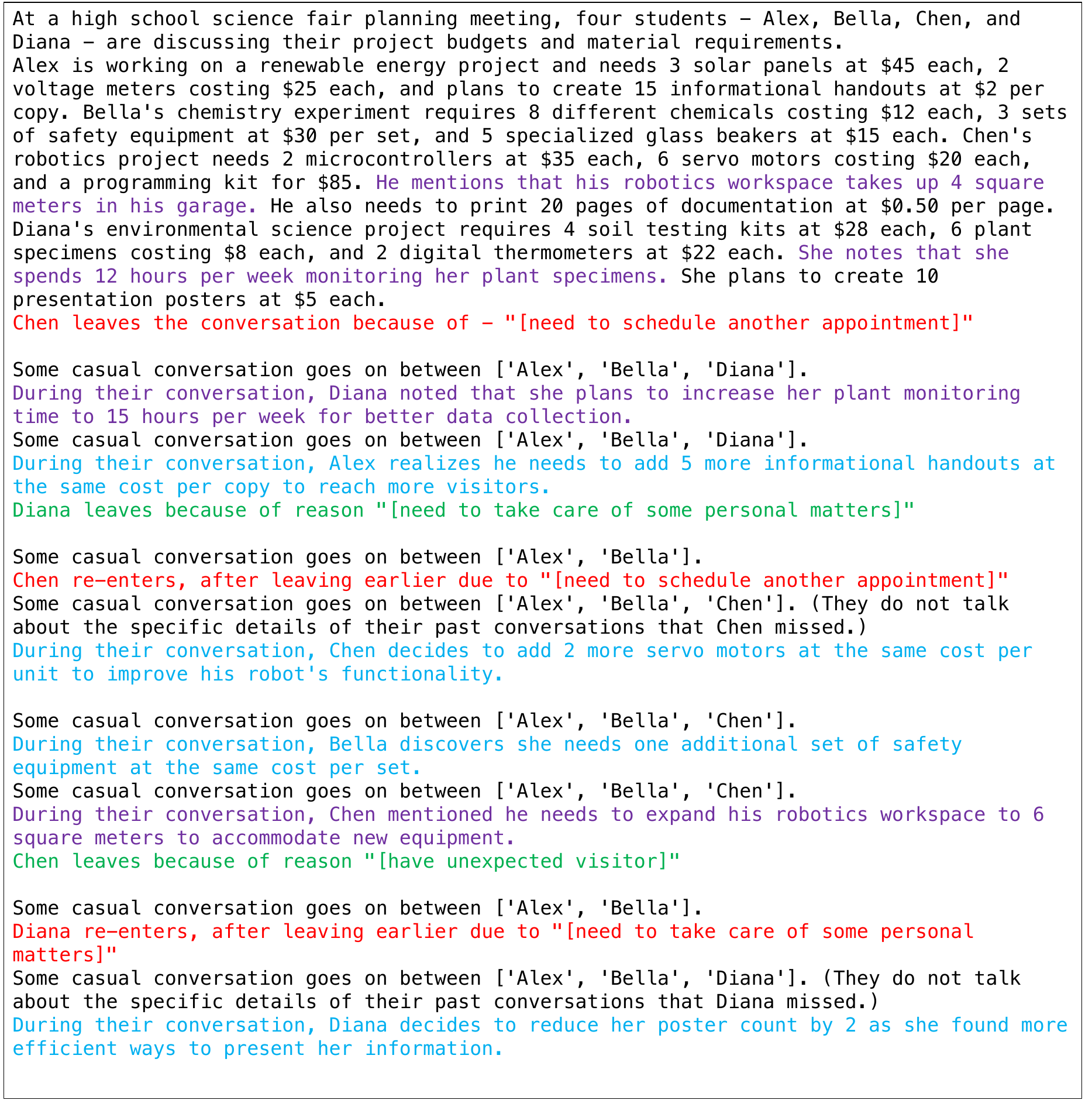}
    \caption{The Distractor Script is created through two components: (1) incorporating distractor variable information into the base premise, and (2) applying variable state perturbations to both (\textcolor{violet}{violet}) distractor variables and (\textcolor{cyan}{cyan}) relevant variables within the premise.}
    \label{Fig:full_script_c33}
\end{figure}

\newpage
\begin{figure*}[h!]
    \centering
    \includegraphics[width=\textwidth]{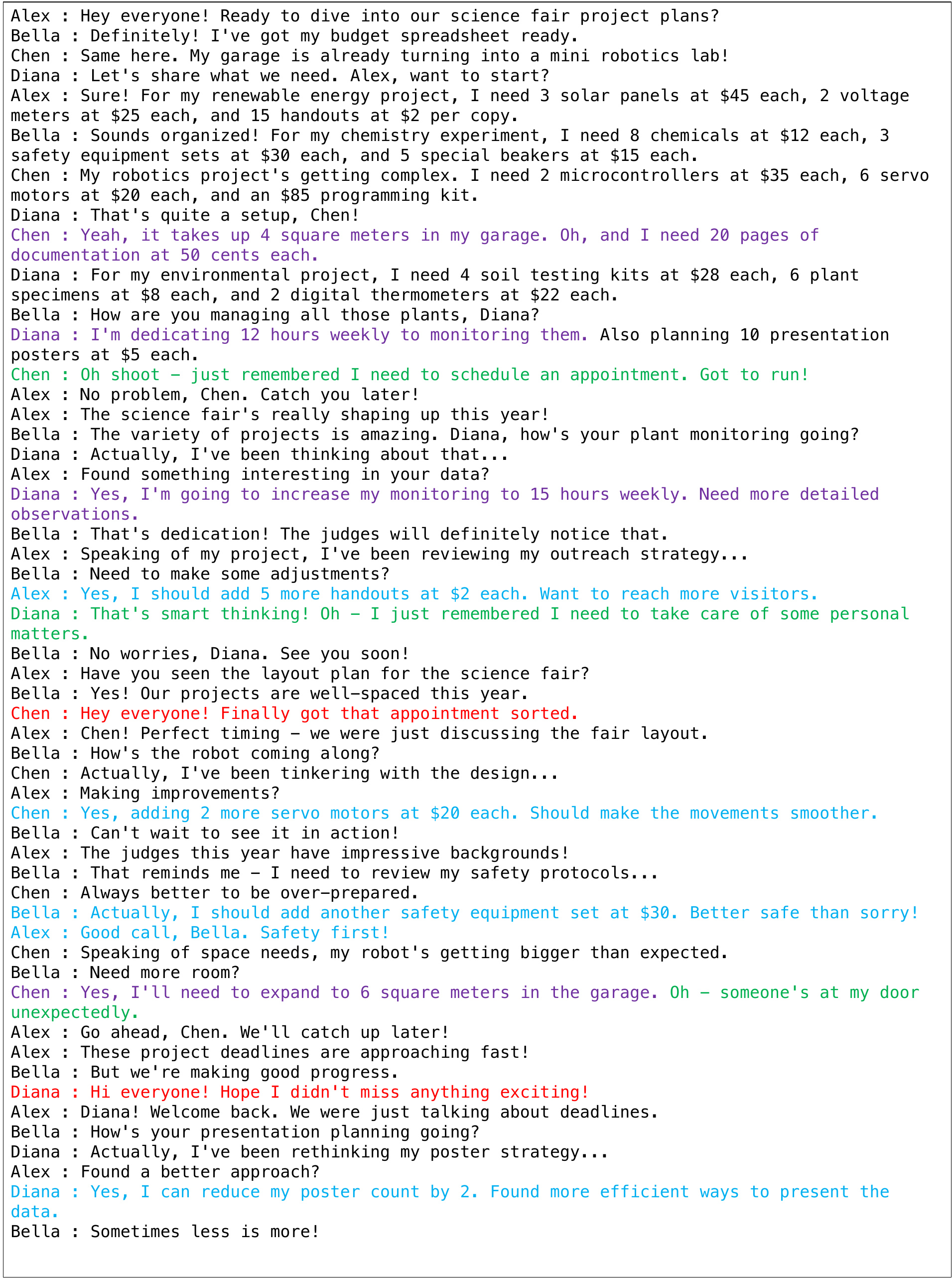}
\caption{Conversation created from distractor script (Figure~\ref{Fig:full_script_c33})}
    \label{Fig:conv_c33}
\end{figure*}

\newpage
\subsubsection{Unanswerable Conversations}
\begin{figure}[h!]
    \centering
    \includegraphics[width=\textwidth]{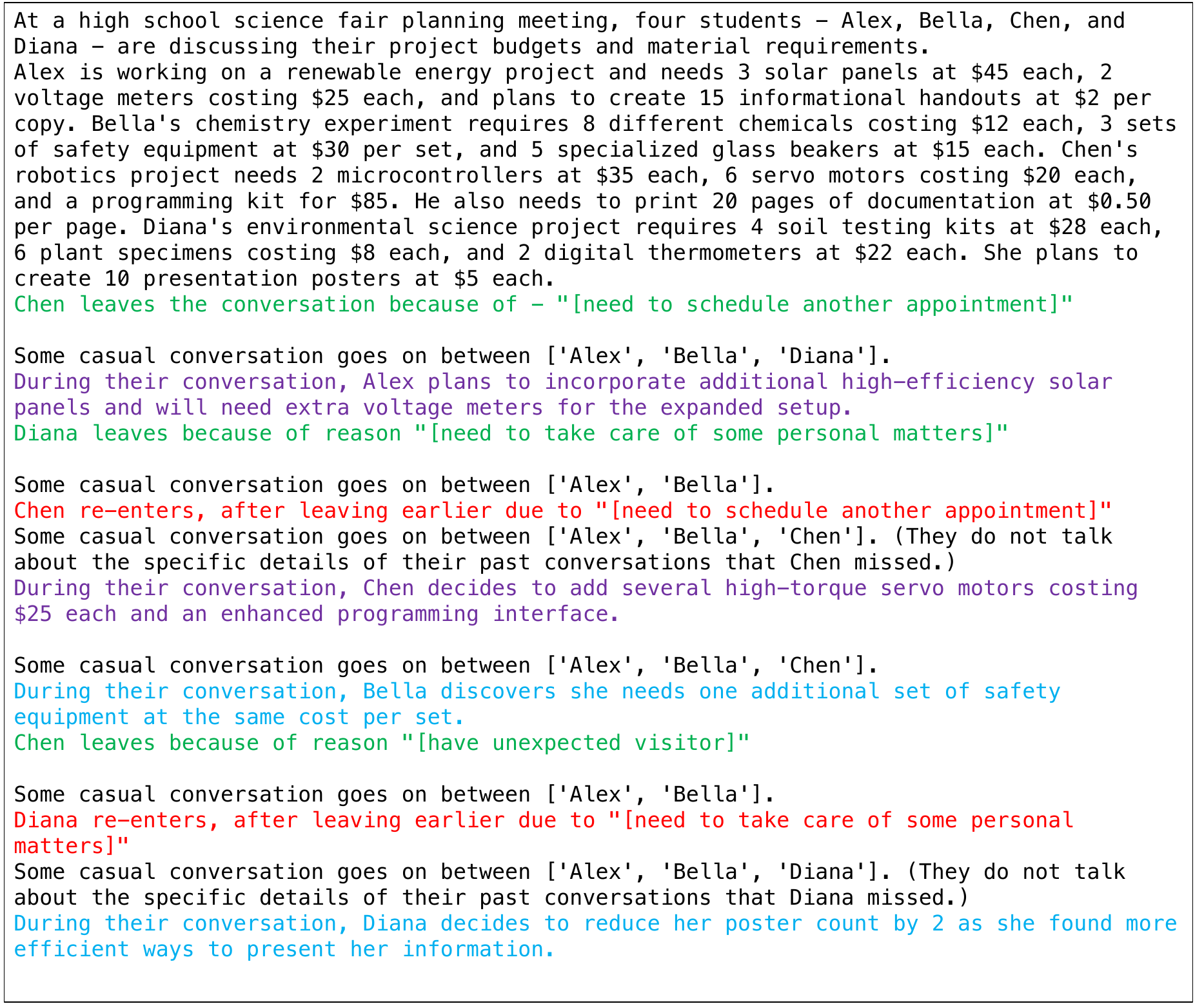}
    \caption{The Unanswerable Script is constructed by combining three elements: (1) the base script premise (Figure~\ref{Fig:premise_dict}), (2) underspecified variable state perturbations (\textcolor{violet}{violet}) (Figure~\ref{Fig:underspec_comps}) for some participants, and (3) sufficiently specified variable state perturbations (\textcolor{cyan}{cyan})(Figure~\ref{Fig:var_change}) for the remaining participants.}
    \label{Fig:full_script_c22}
\end{figure}
\begin{figure*}[h!]
    \centering
    \includegraphics[width=\textwidth]{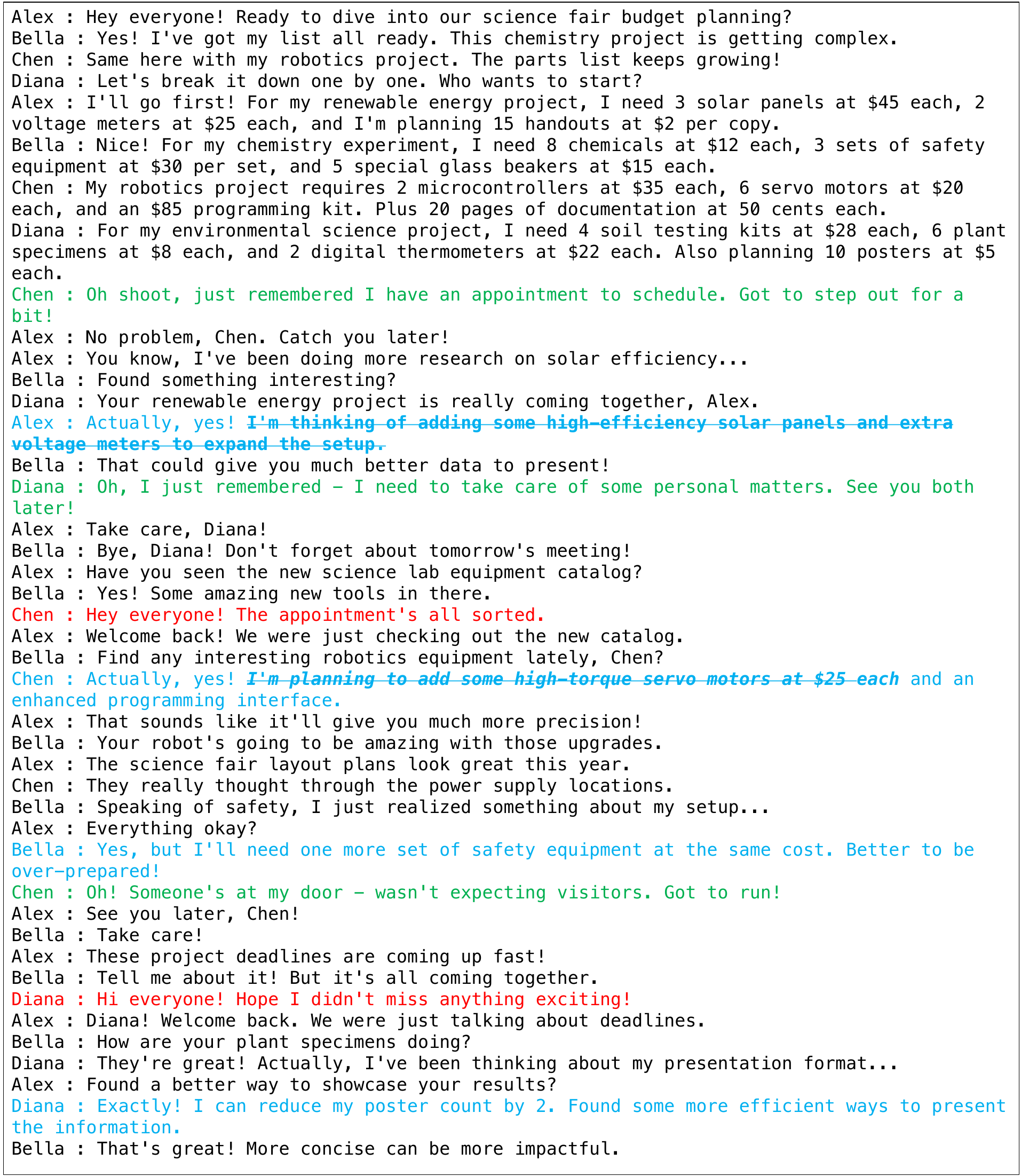}
\caption{Conversation created from an unanswerable script (Figure~\ref{Fig:full_script_c33}). Strike through text shows the underspecificity in the conversation.}
    \label{Fig:conv_c22}
\end{figure*}

\end{document}